\documentclass{article}
\usepackage[numbers]{natbib}
\usepackage{hyperref}
\usepackage{microtype}
\usepackage{graphicx}
\usepackage{subfigure}
\usepackage{placeins} 
\usepackage{booktabs} 
\usepackage{listings} 
\usepackage{courier}
\usepackage{authblk}
\usepackage{geometry}
\usepackage{minitoc}

\usepackage{fullpage}
\usepackage[skip=10pt plus1pt, indent=0pt]{parskip}
\let\svthefootnote\thefootnote
\newcommand\freefootnote[1]{%
  \begin{NoHyper}
  \let\thefootnote\relax%
  \footnotetext{#1}%
  \let\thefootnote\svthefootnote%
  \end{NoHyper}
}
\renewcommand{\cite}[1]{\citep{#1}}
\setcitestyle{authoryear,round,citesep={;},aysep={,},yysep={;}}

\renewcommand\Affilfont{\small}

\makeatletter
\renewcommand\AB@affilsepx{, \protect\Affilfont}
\makeatother

\usepackage{amsmath}
\usepackage{amssymb}
\usepackage{mathtools}
\usepackage{amsthm}

\usepackage[capitalize,noabbrev]{cleveref}

\theoremstyle{plain}

\theoremstyle{definition}

\theoremstyle{remark}

\usepackage[disable,textsize=tiny]{todonotes}

\newcommand{\appendixsection}[1]{
\FloatBarrier 
\clearpage
\section{#1}
}

\DeclarePairedDelimiterX{\infdivx}[2]{(}{)}{%
  #1\;\delimsize\|\;#2
}

\newcommand{\titlevar}{Distinguishing the Knowable from the Unknowable \\ with Language Models}

\title{\vspace{-4em}\rule{\linewidth}{1pt} \\ \vspace{1em}\textbf{\titlevar} \\ \rule{\linewidth}{1pt}}
\author[1]{Gustaf Ahdritz\textbf{*}}
\author[1]{Tian Qin\textbf{*}}

\author[1]{Nikhil Vyas}
\author[1]{Boaz Barak}
\author[1]{\mbox{Benjamin L. Edelman}}
\affil[1]{Harvard University}
\date{}

\makeatletter
\renewenvironment{abstract}{%
    \if@twocolumn
      \section*{\abstractname}%
    \else 
      \begin{center}%
        {\bfseries \abstractname\vspace{\z@}}
      \end{center}%
      \quotation
    \fi}
    {\if@twocolumn\else\endquotation\fi}
\makeatother

\begin{document}
\maketitle
\doparttoc 
\faketableofcontents 

\part{} 

\vspace{-60pt}

\begin{abstract}
\noindent We study the feasibility of identifying \textit{epistemic} uncertainty (reflecting a lack of knowledge), as opposed to \textit{aleatoric} uncertainty (reflecting entropy in the underlying distribution), in the outputs of large language models (LLMs) over free-form text. In the absence of ground-truth probabilities, we explore a setting where, in order to (approximately) disentangle a given LLM's uncertainty, a significantly larger model stands in as a proxy for the ground truth. We show that small linear probes trained on the embeddings of frozen, pretrained models accurately predict when larger models will be more confident at the token level and that probes trained on one text domain generalize to others. Going further, we propose a fully unsupervised method that achieves non-trivial accuracy on the same task. Taken together, we interpret these results as evidence that LLMs naturally contain internal representations of different types of uncertainty that could potentially be leveraged to devise more informative indicators of model confidence in diverse practical settings.
\end{abstract}

\freefootnote{* denotes equal contribution. Correspondence to: Gustaf Ahdritz $\langle$gahdritz@g.harvard.edu$\rangle$, Tian Qin $\langle$tqin@g.harvard.edu$\rangle$.}

\section{Introduction}
\label{submission}

Large language models (LLMs) are remarkably well-calibrated; in question-answering settings, token-level probabilities output by a pre-trained model tend to match the rates at which the corresponding tokens actually occur \cite{anthropic_calibration, gpt4}. Nevertheless, the degree of a model's uncertainty alone is not always informative, since uncertainty can arise from multiple sources.

The token that follows  ``V\"{a}nern is the largest lake in \_\_\_'' is largely deterministic, and so uncertainty on this prompt can be called knowledge-based, or \emph{epistemic}. In most cases, however, the probability distribution of free-form text exhibits inherent, or \emph{aleatoric}, entropy. For example, there are multiple natural continuations for the prompt ``V\"{a}nern is \_\_\_''. Unlike epistemic uncertainty, aleatoric  uncertainty is not \emph{reducible:} it does not disappear even in the limit of infinitely large models that have been trained with infinite data.

In some structured settings---\emph{e.g.} multiple-choice question answering---it can be clear whether there is a single answer in the ground truth and hence how to classify model uncertainty. Indeed, prior work has primarily focused on these cases \citep{epinet, epinet_llm, anthropic_calibration, cole_selective_answering, semantic_uncertainty}. But in general, whether model uncertainty is epistemic or aleatoric (or both at once) is more difficult to determine.

\begin{figure*}[!t]
\begin{center}
\centerline{\includegraphics[width=0.8\textwidth]{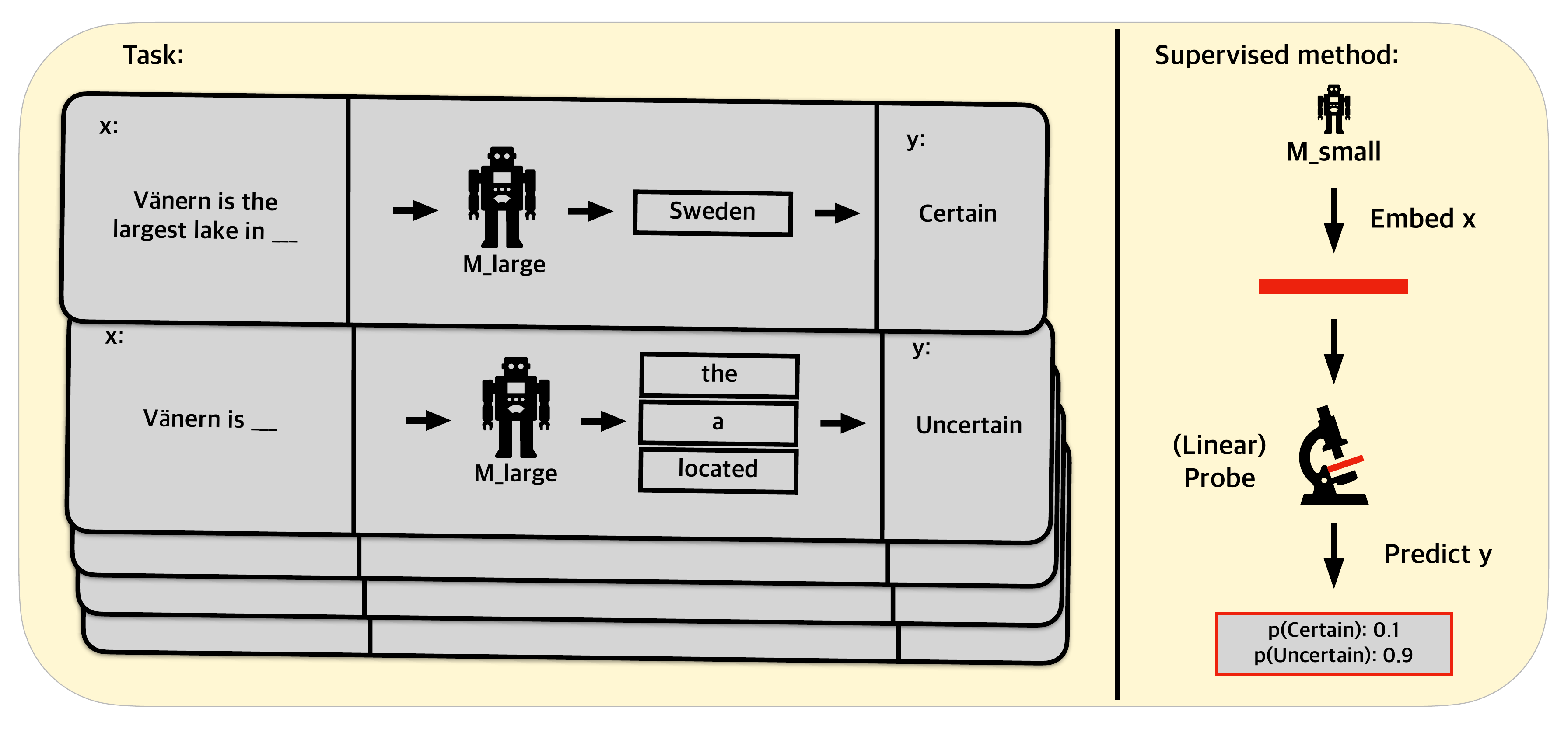}}
\caption{\textbf{Simplified overview of our classification task and the supervised method.} \textit{Left:} The supervised uncertainty classification task. A dataset of labeled prompts is created by running a large LLM on existing text and thresholding its predictive entropy near zero. \textit{Right:} We train probes on activations of a smaller target model to predict the resulting labels without access to the large model. }
\label{overview_figure}
\end{center}

\begin{center}
\centerline{
\includegraphics[width=0.8\textwidth]{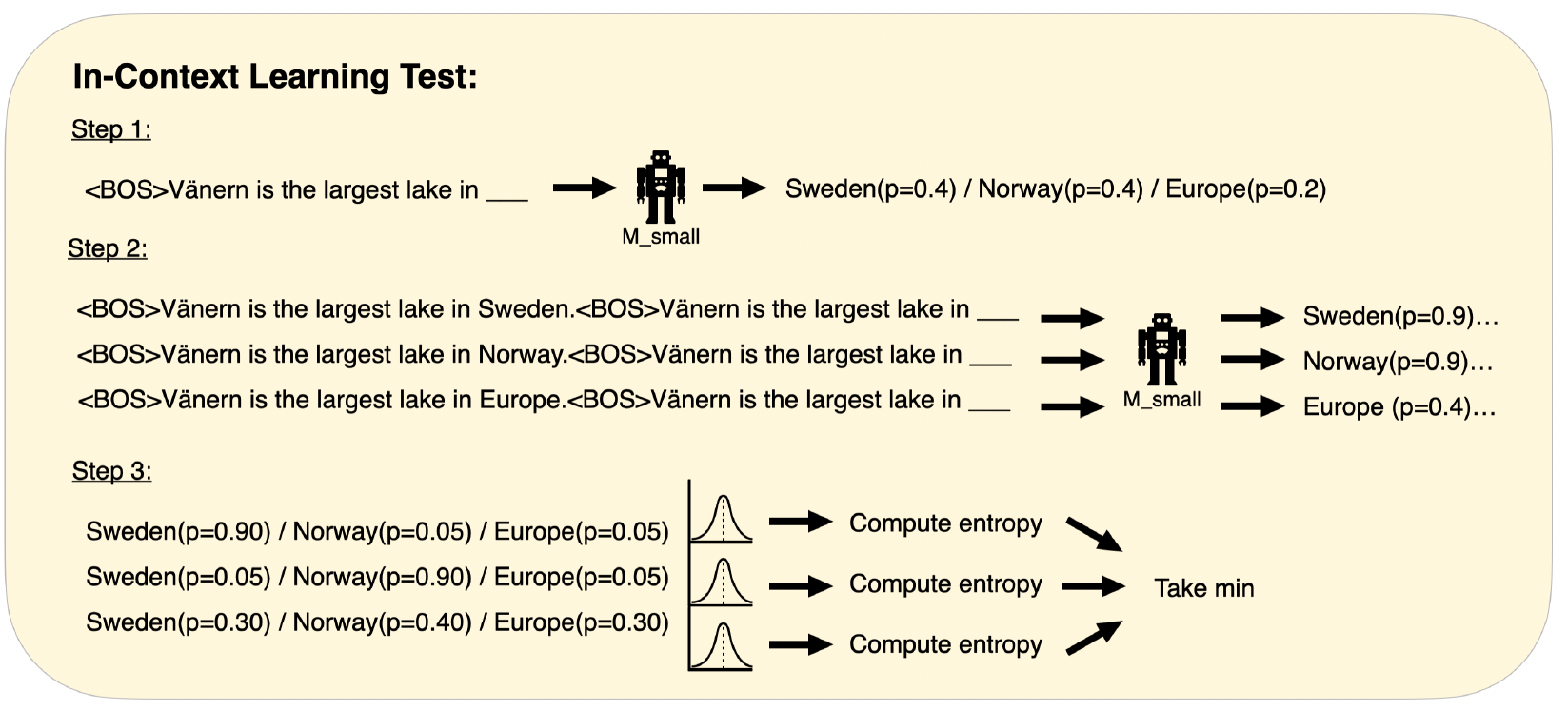}}
\caption{\textbf{The unsupervised ICLT method.} \textit{Left}: Illustration of the unsupervised In-Context Learning Test (ICLT) method. Given a prompt, we first use the small model to produce several next-token candidates. For each candidate, we create a new prompt of the form $\langle\texttt{orig-prompt}\rangle \langle\texttt{candidate}\rangle \langle\texttt{orig-prompt}\rangle$, with document separator tokens before each copy of $\langle\texttt{orig-prompt}\rangle$. We feed these repetition prompts back into the small model, and obtain new next-token distributions. Finally, we use the minimum of the entropies of these distributions as a predictor of the uncertainty of the large model (see the classification task defined in Figure \ref{overview_figure}. In particular, we expect that when the small model's uncertainty is primarily epistemic, it will be more liable to repeat the completion provided in its context, effectively updating on this ``new information''.}
\label{fig:repetition_demo}
\end{center}
\vskip -0.2in
\end{figure*}

In this work, we take the first steps toward a new approach for identifying epistemic uncertainty in completely unconstrained text at the token level. Since the ``ground truth distribution'' is, of course, unknown, we use the assumption---validated by scaling laws \cite{scaling_law, chinchilla, llama}---that larger and more compute-heavy models are better proxies of said distribution and generally exhibit less epistemic uncertainty. Accordingly, we contrast the outputs of small target models with those of the largest models at our disposal to generate token labels. Given a token on which a small target model is uncertain, we run the same prompt through the large model. If the large model is confident about its prediction, we consider the small model's uncertainty about the prediction of this token to be epistemic. Conversely, if the large model is also unconfident about its token prediction, we consider the small model's uncertainty to be approximately ``aleatoric,'' or ``aleatoric-like'' (including tokens about which the large model is itself epistemically uncertain). See Figure \ref{overview_figure} for an illustration of the labeling scheme.

We describe supervised and unsupervised methods for this task. In the supervised setting, we train both linear and non-linear probes on different activations of the small model to predict the uncertainty label on different variations of the task. Figure \ref{overview_figure} (right) contains a simplified overview. The unsupervised method, which we call the \emph{In-Context Learning Test} (ICLT), is inspired by the in-context learning capability of language models. Specifically, we hypothesize that models exhibit different in-context learning behaviors depending on the nature of their uncertainty. See Figure~\ref{fig:repetition_demo} for a depiction of the method.




Our work is ultimately motivated by language model hallucinations \cite{hallucinations, hallucination_survey}. To the extent that hallucinations are a consequence of naively sampling tokens in the presence specifically of epistemic uncertainty,\footnote{This is not the only source of hallucinations, which can also arise from \textit{e.g.} false information in the training set. Nevertheless, we expect that the majority can be described this way.} tagging instances of primarily epistemic uncertainty could be the first step of a new regimen to improve the truthfulness of LLM generations. With a reliable enough uncertainty classifier, one could intervene during generation to avoid tokens where the model's uncertainty is primarily epistemic and the risk of hallucinations is high, or just highlight such tokens in a user interface. Its predictions could also be used to fine-tune models to avoid these tokens themselves, analogously to popular approaches for promoting desirable model behaviors with human feedback \cite{rlhf_ziegler, rlhf_ouyang, dpo}.

For more discussion of our setup and some of its implications, we provide an FAQ in Appendix \ref{faq}.


Our contributions can be summarized as follows: 
\begin{enumerate}
\item We show that small (linear) probes learn to accurately predict (AUC $> 0.9$) when the large model will be confident on individual tokens across model pairings and datasets. See Figure \ref{overview_figure} for details.
\item We show that heads trained to disentangle uncertainty on text from one domain---in our case, Wikipedia articles---transfer to others, like code (AUC $> 0.8$). This suggests that the heads are not simply learning to rely on domain-specific token correlations or other heuristics to achieve high accuracy but may be ``reading'' more robust internal representations of uncertainty present in the small model.
\item We investigate an unrelated, fully unsupervised method for the same task (Figure \ref{fig:repetition_demo}), inspired by in-context learning, and obtain non-trivial results.
\end{enumerate}

\begin{figure*}[!th]
\begin{center}
\centerline{\includegraphics[width=0.7\textwidth]{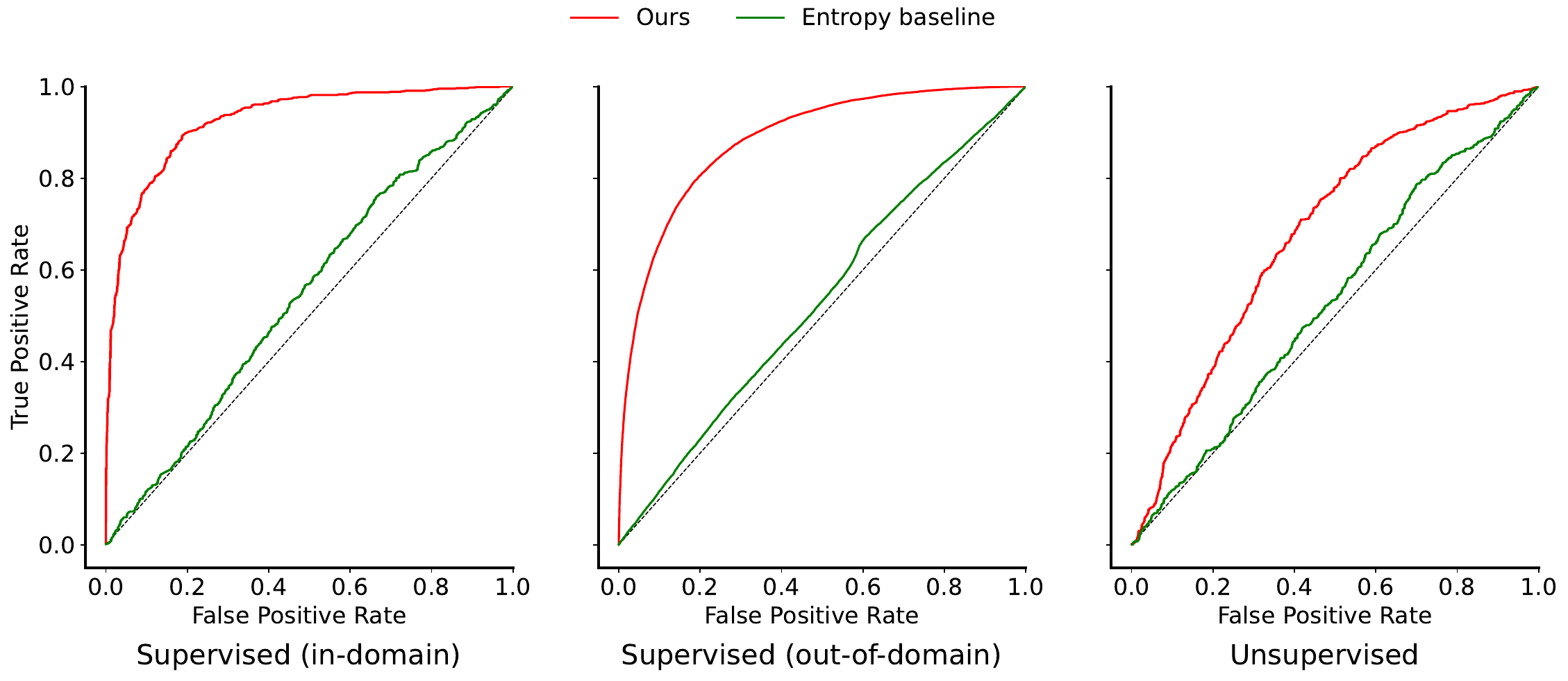}}
\caption{\textbf{High-level classification  results for our supervised and unsupervised methods.} ROC curves for linear probes trained with the model pairing LLaMA 7B / LLaMA 65B, evaluated both in (\textit{left}) and out (\textit{middle}) of the distribution of the probes' Wikipedia training data. All classifier probes are evaluated on balanced test sets. \textit{Right}: ROC curve for unsupervised ICLT method with the model pairing LLaMA 7B / LLaMA 65B evaluated on the  Wikipedia test set.}
\label{overview_roc}
\end{center}
\end{figure*}

\section{Related work}
\label{related_work}

\subsection{Identifying epistemic uncertainty}

Disentangling uncertainty in probability distributions output by neural networks is a longstanding and active area of research \cite{uncertainty_survey}.
Bayesian neural networks (BNNs) maintain (approximate) posteriors of the parameters of a neural network, offering a principled way to reason about model uncertainty, albeit at a prohibitive computational cost \cite{bnns_computationally_intensive}. While ensemble-based approximations of said posteriors have shown some promise \cite{osband_2015, dropout_uncertainty, lakshminarayanan_2017}, scaling these methods to modern LLMs has proven challenging \cite{gleave_2022, neural_testbed}. On the other hand, concurrent work has shown that ensembling clarifications of the inputs for state-of-the-art LLMs---rather than multiple LLMs---provide reliable uncertainty estimates in question-answering settings \cite{hou2023decomposing}.

Epistemic neural networks, or ``epinets,'' are modified neural networks conditioned on an additional epistemic index that produce expressive joint predictions over each combination of classes in a classification task. Changes in the output of an epinet induced by varying the epistemic index can be used to estimate the degree of epistemic uncertainty in the output \cite{epinet}. Small epinets trained on the final representations of ResNets \cite{resnet} and BERT language models \cite{bert} can produce joint predictions for ImageNet \cite{imagenet} and GLUE \cite{glue} classification tasks, respectively. They have also shown promise in active learning, as they can be used to promote epistemically uncertain training examples during model fine-tuning \cite{epinet_llm}. However, they have generally been evaluated in-distribution on relatively small models---at and below 100M parameters---and simple classification tasks, with limited success elsewhere \cite{epinet_llama}.

\citet{anthropic_calibration} directly fine-tunes large language models to predict the probability that they answer well-formed questions correctly, indirectly estimating epistemic uncertainty. The authors achieve high accuracy in-distribution and demonstrate promising trends; confidence predictions on out-of-distribution (OOD) questions are still accurate, larger models are better able to estimate their OOD uncertainty than smaller ones, and confidence predictions tend to increase as relevant ``hints'' are provided in-context. \citet{lin2022teaching} obtains similar results by fine-tuning language models to output confidence on arithmetic tasks in text. Both works study extremely large language models, up to the scale of GPT-3 \cite{gpt3}. However, both focus on the question-answering setting, where there is one, known answer and uncertainty is effectively always epistemic. Rather than gauge how much epistemic uncertainty is present \textit{conditioned on the fact that uncertainty at a token is primarily epistemic}, we seek to identify tokens where model uncertainty is primarily epistemic; to distinguish between tokens that are ``knowable'' and tokens that are not.

\subsection{In-context learning}
While LLMs store knowledge in their parameters \cite{petroni2019language} and can learn new knowledge directly via fine-tuning \cite{de2021editing, mitchell2022memory, mitchell2021fast}, they are also adept at learning in-context \cite{gpt3}. \citet{si_prompting_gpt, zheng2023can, pezeshkpour2023measuring} demonstrate that, when relevant new information is added to prompts, LLMs can update their predictions even when said information is in conflict with model's internal knowledge. \citet{si_prompting_gpt} also shows empirically that larger models are better at updating their knowledge in this way. Our unsupervised method relies on identifying when models rely most heavily on their in-context learning capabilities.

\section{Setup}

\subsection{High-level task description}

Consider a setting where we have access to $M_{\text{small}}$, a comparatively small but still useful language model, and $M_{\text{large}}$, a larger and significantly more capable but impractically expensive counterpart. For convenience, we assume in this work that the two language models share the same vocabulary $\mathcal{T}$, but this is not necessary. 

At a high level, we are interested in the following task. Given a text prompt $x = \{x_i\}_{i = 1}^N$, letting $x_i$ be the $i$-th token of $x$ and $x_{i}^{j}$ be the substring of $x$ between indices $i$ and $j$, we wish to identify indices $k$ where the distributions $M_{\text{small}}(x_{1}^{k})$ and $M_{\text{large}}(x_{1}^{k})$ are substantially dissimilar \textit{without access to $M_{\text{large}}$}. Under the assumption that language models are generally well-calibrated, we are particularly interested in tokens about which the small language model appears uncertain (\textit{i.e.} tokens for which  predictions have large entropy). Note that a successful method for this task would in principle permit a number of useful interventions on $M_{\text{small}}$. For example, if during autoregressive generation using $M_{\text{small}}$ an (accurate) prediction head indicates that $M_{\text{large}}$ is extremely confident about a token on which $M_{\text{small}}$ is uncertain---indicating epistemic uncertainty at that token on the part of $M_{\text{small}}$\footnote{Barring undue training set memorization by the large language model.}---it may be prudent to rewind generation and resample, or simply to highlight the token in a user interface for $M_{\text{small}}$. While $M_{\text{large}}$ may not itself be perfect, improving the capabilities of $M_{\text{large}}$ can also be expected to improve the utility of such a head.

\begin{figure*}[!t]
\vskip 0.2in
\begin{center}
\includegraphics[width=0.95\textwidth]{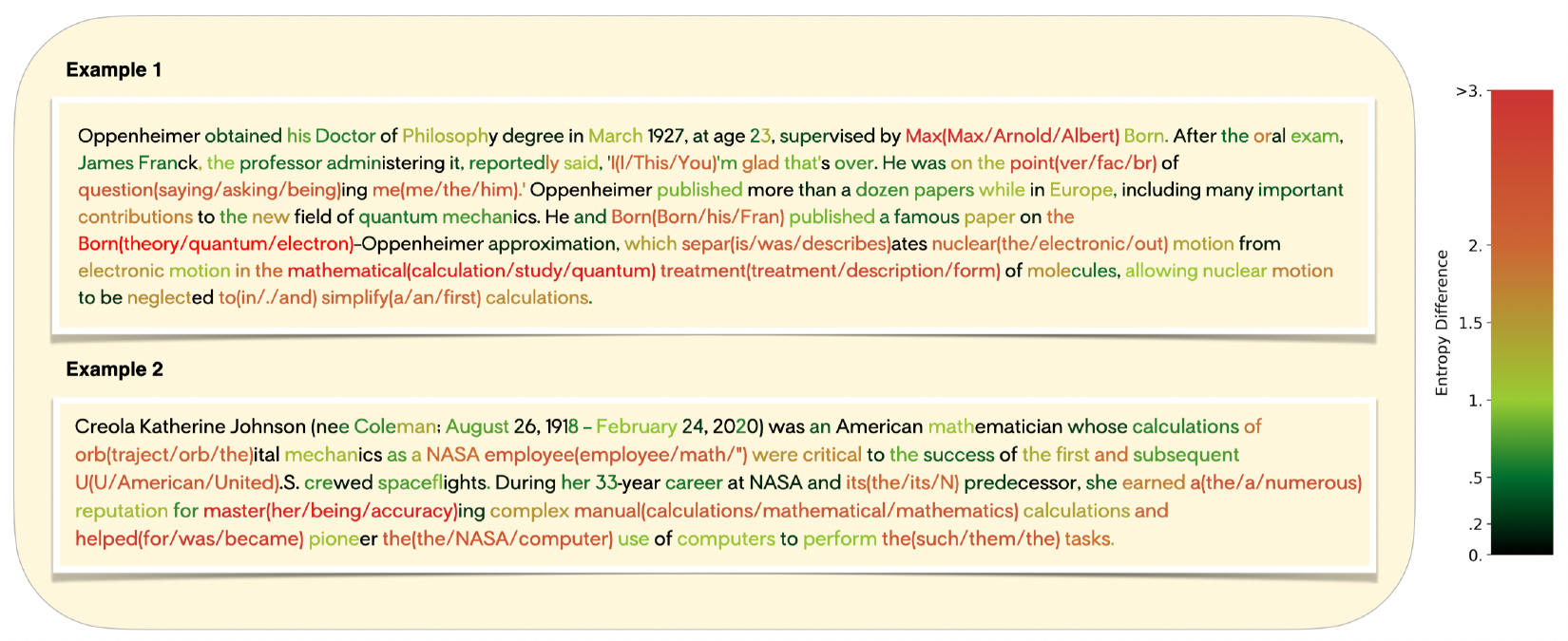}

\caption{\textbf{Given a knowledgeable enough ``large'' model, tokens where a small model is unconfident while the large model is confident can be thought of as instances of epistemic uncertainty on the part of the small model.} Snippets from Wikipedia articles with tokens highlighted wherever the conditional predictive entropies of a pair of small and large language models (LLaMAs 7B and 30B) differ. For tokens where the entropy difference is larger than 2 bits, we also display the small model's top 3 token predictions in parentheses. Qualitatively, we observe that tokens at which the large model has near-zero entropy but the small one does not tend to be ``epistemic'' in nature, corresponding to \textit{e.g.} dates, people, and specific technical vocabulary. See Appendix \ref{disagreements_autoregressive} for an autoregressive version of the same.}
\label{disagreements}
\end{center}
\end{figure*}

We have not yet defined ``substantially dissimilar.'' The most granular (and practically applicable) version of this task would involve predicting the difference in probability between $M_{\text{small}}$ and $M_{\text{large}}$ at each token in $\mathcal{T}$, but for now we focus on a simpler variant: predicting high-level summary statistics about $M_{\text{large}}$'s output distribution using only intermediate representations from $M_{\text{small}}$. We experiment with a variety of closely related target values (see Appendix \ref{training_objectives} for details), but unless otherwise noted we default to the conceptually simplest option: the Shannon entropy of the large model's prediction:
\begin{equation*}
        H(M_{\text{large}}(x)) = -\sum_{t \in \mathcal{T}} M_{\text{large}}(x)_t \log{M_{\text{large}}(x)_t}.
    \end{equation*}

A low value indicates that the large model is placing probability mass on a small number of tokens, indicating the large model is fairly certain about its next token prediction. In this case, we consider that the small model's uncertainty is ``epistemic-like." On the other hand, a high value indicates that the large model is also uncertain and we consider the small model's uncertainty to be ``aleatoric-like." In Figure \ref{disagreements}, we visualize differences in paired model entropies, and in Appendix \ref{labeled_examples}, we provide examples of labeled tokens in real data.

There are clear limitations to our framing, most importantly that our  ``large'' language models still exhibit epistemic uncertainty in their own right, which introduces label noise. We also ignore sequence-level, semantic uncertainty \cite{semantic_uncertainty} and mixtures of epistemic and aleatoric uncertainty.\footnote{Granted, it could be argued that it is often possible to pinpoint one token where a model ultimately commits to an output it is  conceptually uncertain about.} Nevertheless, we believe that solving this narrower but still nontrivial problem is a meaningful first step.

%

\subsection{Models}

While we stress again that our task setup does not necessarily require that both the small and large models share the same vocabulary, for convenience, we consider ``hierarchies'' of open-source language models like LLaMA \cite{llama}, Llama 2 (chat and non-chat variants) \cite{llama_2}, and Pythia \cite{pythia}. Within each family, vocabulary, architecture, and training data are shared, allowing for simple evaluation across model sizes and, in the case of Pythia, training checkpoints. LLaMA models are available at 7B, 13B, 33B, and 65B parameters, Llama 2 at 7B, 13B, and 70B (with and without instruction tuning), and Pythia at 70M, 160M, 410M, 1B, 1.4B, 2.8B, 6.9B, and 12B (along with intermediate checkpoints for each). 

\subsection{Datasets}
\label{datasets}

For all experiments, we use free-form text data not in the training data of the small and large models in each pair. It is specifically important that the large language model not have encountered and memorized the text in question; otherwise, we wouldn't expect to be able to predict the large model's confidence without specific knowledge of its training set. Ideally, because we are most interested in small model uncertainty attributable to content rather than just text format, we also prefer data in a format familiar to both models. For LLaMA models, we use the set of Wikipedia articles created (\textit{not} last edited) between the models' training cutoff and June 2023. Older Wikipedia data is present in the LLaMa models’ training set \cite{llama, llama_2}. We also use the designated Pile evaluation and test sets \cite{pile}. Each Pile subset can be further subdivided into the Pile's component parts. In our out-of-distribution evaluations, we focus on three: ``Code,'' containing code data from GitHub, ``Stack Exchange,'' data from Q\&A forums, and ``EuroParl,'' a smaller subset of multilingual European Parliament proceedings.\footnote{Corresponding to the \texttt{pile\_set\_name} \texttt{Github}, \texttt{StackExchange}, and \texttt{EuroParl}, respectively.} For samples from each set, see Section \ref{dataset_samples} in the appendix.

\begin{figure}[!t]
\vskip 0.2in
\begin{center}
\centerline{\includegraphics[width=\columnwidth]{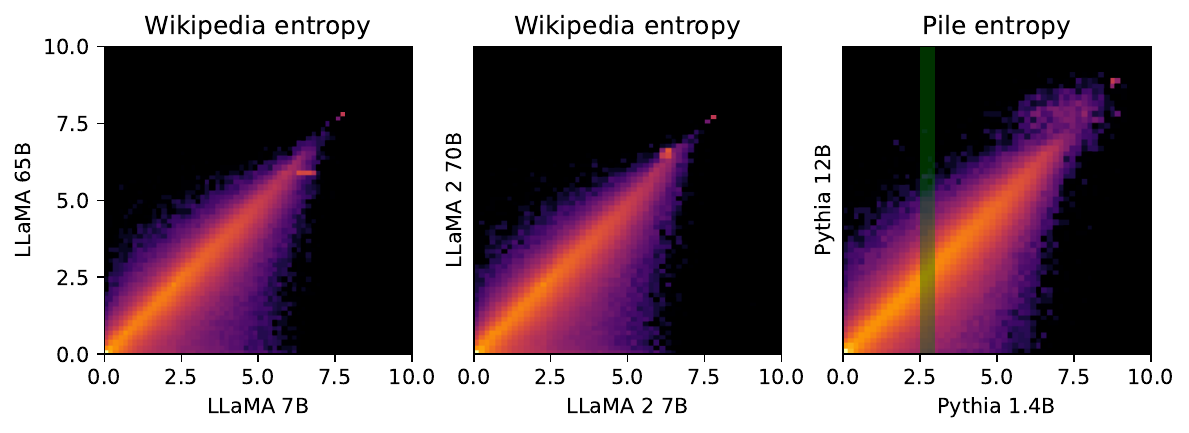}}
\caption{\textbf{Small model entropy and large model entropy are heavily correlated.} Heatmaps (with log-scale color schemes) of the entropies of token probabilities output by smaller and larger language models. Across model types and datasets, these values are heavily correlated. To rule out that our heads learn to rely on this fact, we train separate heads on tokens from narrow bands of the small model's predictive entropy (example band highlighted in green) within which the correlation is negligible.}
\label{entropy_is_correlated}
\end{center}
\end{figure}

\subsection{Baselines}
\label{baselines}

Correlations between the predictions of the large and small models make it possible to predict the entropy of the large model's prediction simply by outputting the entropy of the small model's prediction (see Figure \ref{entropy_is_correlated}). Similarly, certain tokens are correlated with high- or low-entropy predictions for the following token. Any model will often have high entropy after a period, for example. While we filter our data to mitigate these effects and ensure that our heads are learning nontrivial functions, as sanity checks, we also include the following baseline methods wherever relevant:

\textbf{Best entropy threshold (BET):} For our binary classification tasks, we list the accuracy of the best possible classifier that outputs labels based on a threshold of the small model's entropy.

\textbf{Best entropy threshold after finetuning (BET-FT):} To separate out the effects of the fact that our heads are trained on data that is slightly out of the corresponding language model's training distribution, we finetune the small language model's final language modeling layer on distributions predicted by the large model and repeat the BET benchmark.

\textbf{Small model entropy (SME):} Analogously, for our regression tasks, we list the error of simply predicting the entropy of the small model's predicted distribution.

\textbf{Prediction from initial embeddings (PIE):} As an input to our classifier and regression heads, we replace the final embedding output by the small model with the initial embedding in the small model, before any transformer layers. This can only be effective if it is possible to predict the large model's confidence from the preceding token alone.

\section{Supervised experiments}

\subsection{Training details}
\label{training_details}

To train all following heads we use Adam \cite{adam} and a learning rate of $10^{-5}$. Heads have a hidden dimension of 2048 and either one or zero hidden layers (in the nonlinear and linear cases, respectively). Classification heads are trained with standard cross-entropy loss; regression heads with least squares. All heads are trained with early stopping based on validation loss. We use PyTorch \cite{pytorch} and A100 GPUs.\footnote{Code for all experiments is available here: \url{https://github.com/gahdritz/llm_uncertainty}}

\subsection{Binary classification (with gap)}
\label{classification_with_gap}
\textbf{Setup:} We begin with the ``easiest'' variant of the task: identifying tokens for which the large model's entropy is close to zero.

As previously noted, an important consideration in our experiments is that the entropies of the predicted distributions from the small model and the large model tend to be heavily correlated, meaning that it is possible to trivially perform well at tasks depending on predicting the entropy of the large model's distribution simply by \textit{e.g.} computing the entropy of the small model's prediction directly. For that reason, we try training separate classifiers for different ``bands,'' or ranges of the values of the small model's prediction's entropy (\textit{e.g.} one dedicated classifier for all tokens where the small model's next token prediction has entropy between 2 and 3). The narrower the bands, the weaker the aforementioned correlation within each band.

We train both unconditional classification heads and specialized heads for individual bands. Heads take as input next token embeddings from $S$. We find embeddings from the middle layers of each model are best (see Appendix \ref{choice_of_embedding} for more details), though different layers do sometimes perform better out of distribution. To eliminate error caused by ambiguous tokens near bin boundaries and establish a clearer proof of concept, we initially focus on a binary classification task where we introduce an artificial gap between bins, reducing the task to predicting whether, conditioned on the small model's prediction having entropy inside a narrow band (\textit{e.g.} $[k, k+1]$ for some $k$), the large model predictive entropy is very low or high, \textit{i.e.} close to zero ($\lessapprox 0.2$) or within the same narrow band. For precise details on the binary classification setup, see Appendix \ref{binary_classification_setup_gap}. For all classification tasks, we heavily filter tokens to 1) balance class labels and 2) equalize counts of the token immediately before the target token across classes to eliminate trivial correlations. Note that both interventions make the task more difficult.

\textbf{Transfer to unseen distributions:} Including this filtering process, we have taken several measures to prevent our heads from learning overly simplistic functions. While we believe these measures are sufficient to elicit non-trivial behavior in practice, they do not rule out that our heads learn to depend on heuristic, domain-specific cues, like correlations between class labels and \textit{2}-grams in the prompt, rather than generalizable information computed internally by the small model.\footnote{One could imagine an internal ``flag,'' loosely defined, denoting the presence of epistemic uncertainty.} To determine whether this occurs, we perform additional evaluations on out-of-distribution data. For example, we evaluate heads trained exclusively on tokens from Wikipedia articles on code data from the Pile.

Selected results for the binary variant are given in Table \ref{tab:binary_classification_gap_results}. Additional experiments, including results for different entropy bands and model pairings, are described in Appendix \ref{binary_classification_gap_appendix}. It is important to note that, because class labels are determined by the large model and we balance training and evaluation sets by class, each classifier is trained and evaluated on slightly different subsets of the tokens in its respective dataset. 

In general, with minimal hyperparameter tuning, it is possible to train accurate linear classifiers (AUC $> 0.9$) across model pairings and datasets. These classifiers perform nearly as well outside of their training distribution, on code, multilingual, and Q\&A-style evaluations.

\begin{table}[!t]
\caption{\textbf{Linear classifiers of small model activations can reliably predict when the large model is confident, both in and out of distribution.} AUROC of binary classifiers for large model predictive entropy (with an artificial gap) all trained on the Wikipedia set. Inputs include tokens for which the small model's predictive entropy is in the range [2, 3). Labels correspond to whether the large model's predictive entropy is 1) near zero or 2) within the same band. Training and test sets are all class- and token-balanced to mitigate trivial entropy correlations.}
\label{tab:binary_classification_gap_results}
\vskip 0.15in
\begin{center}
\begin{small}
\begin{sc}
\begin{tabular}{ccccc|cc}
\toprule
     Model & S & L & Type & Test set & AUC & Acc \\
    \midrule
     LLaMA & 7B & 30B & MLP & Wikipedia &  $\mathbf{0.94}$ & 0.87 \\
     LLaMA & 7B & 30B & Linear & Wikipedia &  $\mathbf{0.94}$ & 0.86 \\
     LLaMA & 7B & 30B & BET & Wikipedia &  $0.54$ & 0.54 \\
     LLaMA & 7B & 30B & BET-FT & Wikipedia &  $0.63$ & 0.67 \\
     \midrule
     LLaMA & 7B & 65B & MLP & Wikipedia &  $\mathbf{0.93}$ & 0.86 \\
     LLaMA & 7B & 65B & Linear & Wikipedia &  $\mathbf{0.93}$ & 0.85 \\
     LLaMA & 7B & 65B & BET & Wikipedia &  $0.54$ & 0.55 \\
     LLaMA & 7B & 65B & BET-FT & Wikipedia &  $0.66$ & 0.67 \\
     \midrule
     Pythia & 1.4B & 12B & MLP & Wikipedia &  $\mathbf{0.90}$ & 0.81 \\
     Pythia & 1.4B & 12B & Linear & Wikipedia &  $\mathbf{0.87}$ & 0.79 \\
     Pythia & 1.4B & 12B & BET & Wikipedia &  $0.59$ & 0.59 \\
     Pythia & 1.4B & 12B & BET-FT & Wikipedia &  $0.75$ & 0.71 \\
     \midrule 
     LLaMA & 7B & 30B & Linear & Code &  $\mathbf{0.82}$ & 0.75 \\
     LLaMA & 7B & 30B & Linear & Europarl &  $\mathbf{0.79}$ & 0.71 \\
     LLaMA & 7B & 30B & Linear & Stack Ex. &  $\mathbf{0.88}$ & 0.80 \\
     \midrule
     LLaMA & 7B & 65B & Linear & Code &  $\mathbf{0.79}$ & 0.72 \\
     LLaMA & 7B & 65B & Linear & Europarl &  $\mathbf{0.81}$ & 0.70 \\
     LLaMA & 7B & 65B & Linear & Stack Ex. &  $\mathbf{0.88}$ & 0.80 \\
     \midrule
     Pythia & 1.4B & 12B & Linear & Code &  $\mathbf{0.67}$ & 0.62 \\
     Pythia & 1.4B & 12B & Linear & Europarl &  $\mathbf{0.76}$ & 0.65 \\
     Pythia & 1.4B & 12B & Linear & Stack Ex. &  $\mathbf{0.80}$ & 0.71 \\
\bottomrule
\end{tabular}
\end{sc}
\end{small}
\end{center}
\vskip -0.1in
\end{table}

\subsection{Binary classification (without gap)}
\label{binary_classification_no_gap}

The results in the previous section clearly demonstrate that embeddings from the small model contain enough information to distinguish between tokens where the predicted distribution output by the large model has high or near-zero entropy, but the gap introduced between bins means the resulting classifiers cannot be used in practice. Here, we train binary classifiers without a gap. As before, we use entropy bands and balance classes with equalized previous token counts. We set the boundary between bins to 1 bit (somewhat surprisingly, the choice of threshold does not meaningfully affect the performance of the classifiers). Full results are provided in the appendix in Table \ref{binary_classification_no_gap}.

As expected, points near the boundary cause the accuracy to drop---to \textit{e.g.} approximately 75\% for the 7B/65B LLaMA experiment in the [2, 3) band, down from the high 80s---but the classifiers still outperform both baselines by large margins. We experiment with the more difficult but more flexible approach of training regressions to predict the large model's predictive entropy directly in Section \ref{regression_section} of the appendix.


\section{Unsupervised experiments}

Training our previous methods depends on access to a larger language model. Of course, the larger model is only a proxy for the true distribution of the training data, and one may not always have access to such a model. We have already demonstrated in Section \ref{classification_with_gap} that small (linear) heads can be trained on the small model's embeddings to classify whether the larger model's predictive entropy would be very low or high. In this section, we attempt to elicit such information from the small model directly, and  without any additional training.

Given a prompt $p$, we generate variations of the prompt for in-context learning using the small model's top $k$ token predictions $t_i$, with $i \in [1, k]$. Specifically, separately for each $i\in[1, k]$, we prepend ``$p + t_i$'' to the original prompt. We then feed this ``repeated prompt'' back into the model for next-token prediction and measure the degree to which the model repeats information in its provided ``hint.'' Note that the resulting prompts are repetitive and often highly nonsensical; we do not even complete words in cases where the ``hint'' token is merely a word fragment. Empirically, our method is not sensitive to our choice of how the relevant context is provided (see Appendix \ref{appdx:repetition_context} for an ablation study). See Figure \ref{fig:repetition_demo} for an illustration of the method, which we call the In-Context Learning Test (ICLT). 

The intuition behind our unsupervised method comes from in-context learning. LLMs have demonstrated remarkable capabilities to learn from context \cite{gpt3}. However, existing work typically focuses on using model's in-context learning capabilities to extract latent knowledge from LLMs \cite{petroni2019language, ccs} or instill new knowledge during inference time \cite{si_prompting_gpt, zheng2023can, pezeshkpour2023measuring}. Here, we hypothesize that LLMs learn differently from their contexts in the presence of different types of uncertainty. Specifically, we speculate that they are more likely to simply copy information in their contexts if they are epistemically uncertain and less likely when their uncertainty is primarily aleatoric.

We first design a toy experiment to demonstrate that such ``selective in-context learning'' capabilities are possible for transformer architectures at all. We then try out our unsupervised method on real-world LLMs.
\subsection{Synthetic proof of concept} 
\textbf{Setup:} To build intuition for our method, we first consider a simplified synthetic task. We construct $\langle \text{question}, \text{answer}\rangle$ pairs where each question consists of 1) a single bit indicating whether it's epistemic (0) or aleatoric (1) and 2) bits uniquely identifying the question. The answer is a single bit. For ``epistemic'' questions, answers are drawn uniformly at random once at initialization and then fixed (and can therefore be memorized). For ``aleatoric'' questions, answers are resampled uniformly at random every time the question is encountered. A sample question/answer pair is given below:
\begin{equation*}
\langle \underbrace{0}_{\textit{epistemic}}\underbrace{00101011101011}_{\textit{question index}} \underbrace{1}_{\textit{answer}}\rangle
\end{equation*}
We train a small ($\sim$100M-parameter) language model to predict the answer to questions in the $k$-shot setting, choosing hyperparameters such that target questions are occasionally duplicated as examples in the model's prompt, permitting ``copying'' (details in Appendix \ref{unsupervised_appendix}). We train the model until convergence and examine the model's in-context learning behavior on both types of questions. We expect the model to learn that the nature of its uncertainty for any given question is uniquely determined by the first bit. As a result, the model should perform in-context learning and update its predictions for epistemic questions when prompted with applicable information in the same context. On the other hand, the model should not copy information from its prompt if the question is aleatoric and the model's predicted distribution over the answer choices is correct. 

\begin{figure}[!ht]
\begin{center}
\centerline{\includegraphics[width=0.7\columnwidth]{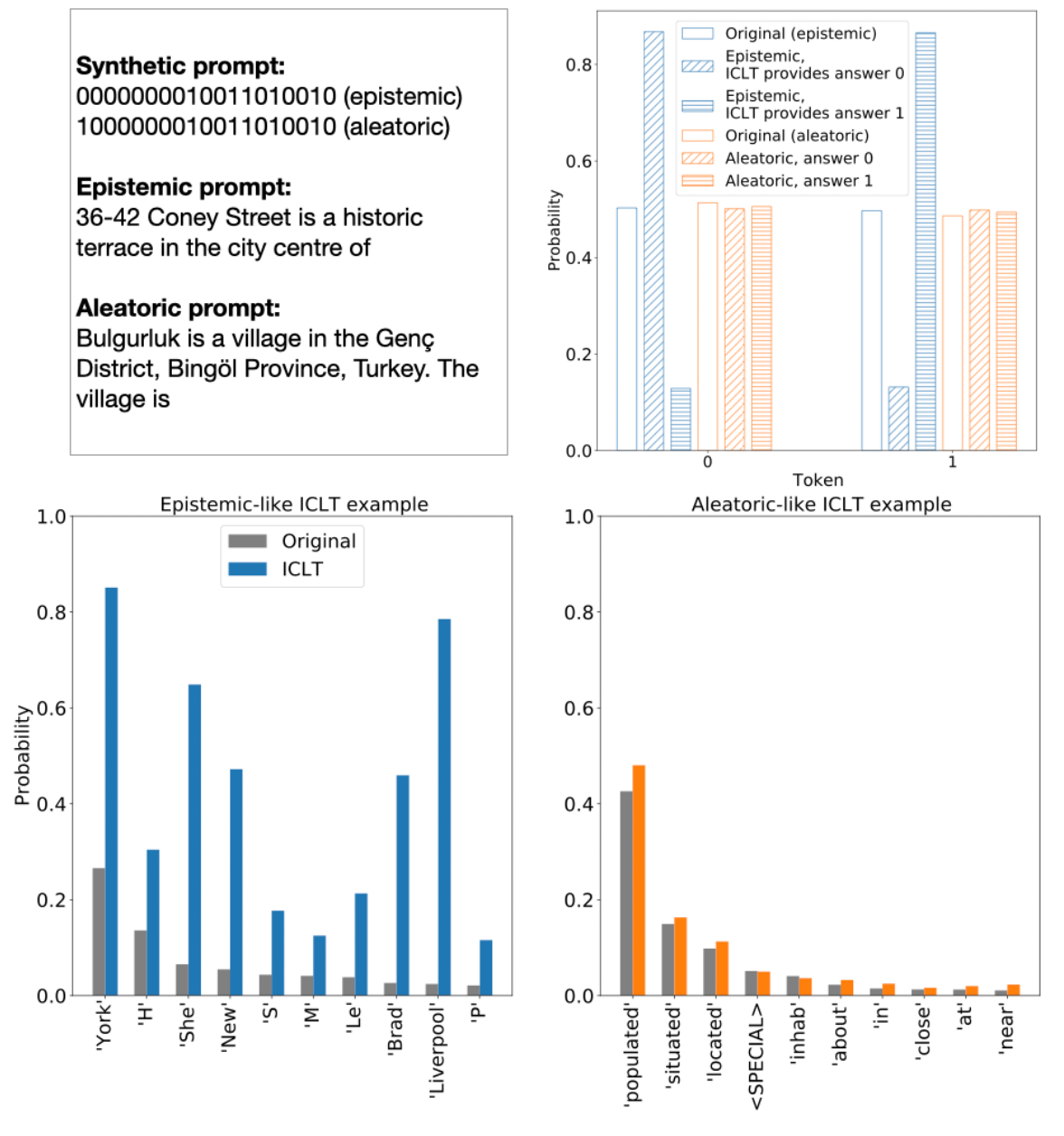}
}
\vspace{-.15in}
\caption{\textbf{Concrete examples of ICLT in synthetic and empirical settings.} \textit{Top right:} Synthetic results. \textit{Bottom:} Empirical results using LLaMA 7B. Gray bars represent the model's original prediction without additional in-context information. Blue and orange bars are the model's predicted probability for token $i$ conditioned on a repeated prompt containing token $i$.}
\label{fig:reptition_synetheic}
\end{center}
\vspace{-.2in}
\end{figure}

\textbf{Results:} In Figure \ref{fig:reptition_synetheic}, upper-right panel, we showcase model behavior for two example questions, one epistemic and one aleatoric. We first prompt the model with the question without any additional context. In both cases, the model initially places approximately the same probability on both answers.  We then prepend a copy of the question and an answer bit to the original prompt and re-run the model to test whether it increases the likelihood of the provided answer. As expected, we observe that the model consistently performs in-context learning to answer epistemic questions, regardless of the correctness of the provided answer, but does not change its predictions in the aleatoric case. Similar knowledge-updating behavior is pointed out by \cite{si_prompting_gpt}. 

\subsection{ICLT on real data}

In the synthetic experiment, to form an internal representation of uncertainty, the model only needs to observe that the epistemic/aleatoric nature of its uncertainty is uniquely determined by the first bit in the question. Real language is obviously less clear-cut. However, success in the synthetic setup hints  that language models can form internal representations of different types of uncertainty and adjust their behavior accordingly. If real language models are capable of the analogous task of classifying natural-language prompts (even in a ``soft'' way), we may be able to extract that information in a similar fashion: by simply measuring how ``suggestible'' the language models are under each prompt.
\begin{figure}[!ht]
\begin{center}
\centerline{\includegraphics[width=0.48\columnwidth]{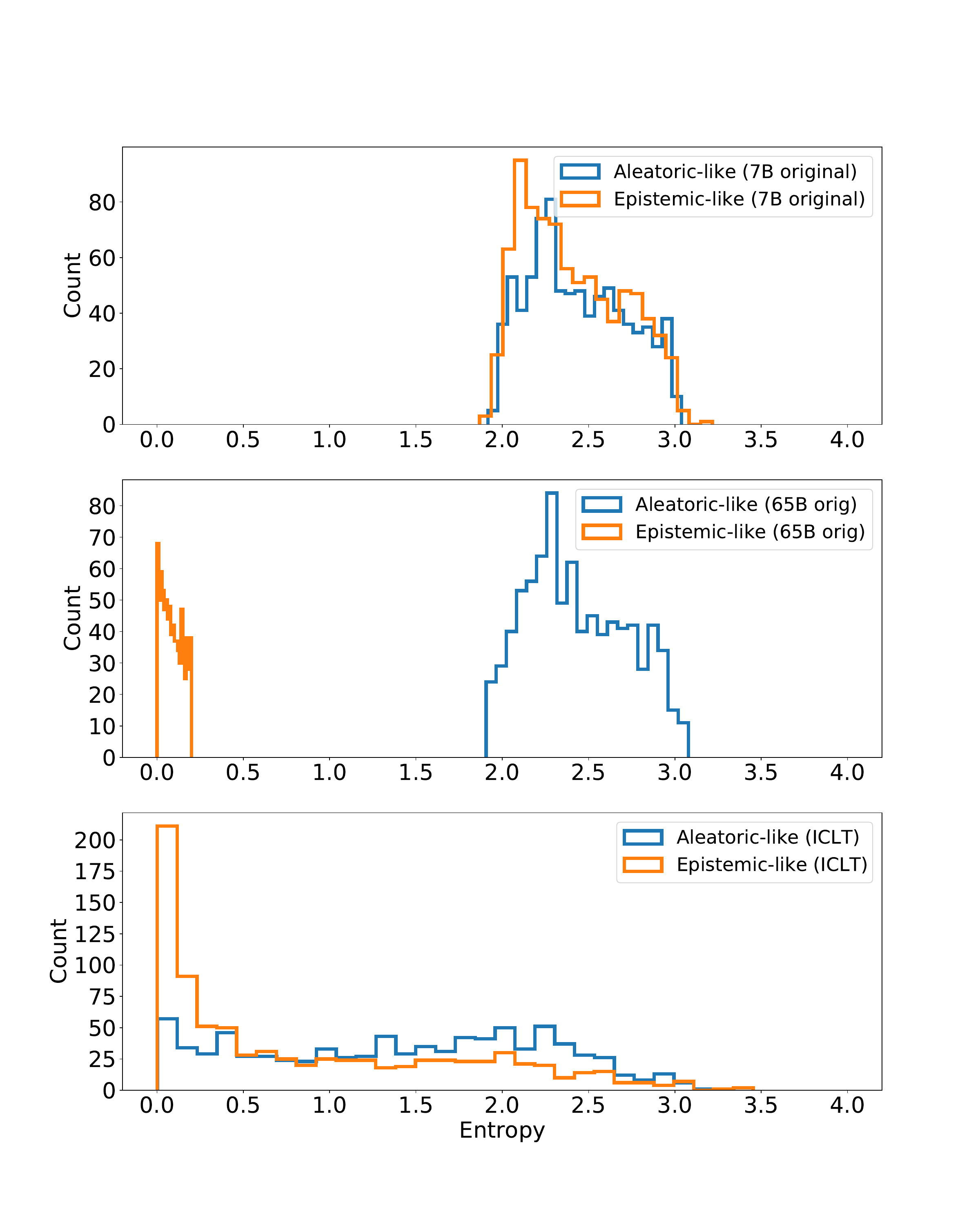}}
\vskip -0.2in
\caption{\textbf{ICLT separates epistemic from aleatoric tokens.} \textit{Top:} Original predictive entropy of LLaMA 7B for tokens in the class-balanced Wikipedia test set (used for SME baseline). \textit{Middle:} Predictive entropy of the 65B model (source of aleatoric/epistemic labels). \textit{Bottom:} Minimum predictive entropy of LLaMA 7B during ICLT. Relative to aleatoric ones, entropy tends to decrease on epistemic examples when a hint is provided in context.}
\label{fig:reptition_entropy}
\end{center}
\vskip -0.2in
\end{figure}
To quantify the result, we look at whether the model significantly increases the likelihood of repeating the token $t_i$ when it is presented in the context compared to its likelihood in the original generation. In the lower two panels of Figure \ref{fig:reptition_synetheic}, we showcase two examples (one epistemic and one aleatoric). The observed behavior agrees with our intuition for ICLT. Note that these are cherry-picked; in Appendix \ref{appdx:rep_additional_examples} we list more diverse examples and speculate why ICLT works less well in some cases.

For a larger-scale evaluation, we employ a larger model to identify epistemic uncertainty, as in Section \ref{classification_with_gap}. We perform ICLT with top $k$ tokens as context and use the minimum entropy produced by the model when prompted with the top $k$ tokens as additional context. We formulate the binary classification task as before, with token-based filtering and a gap between the two bins. Figure \ref{fig:reptition_entropy} shows the original entropy for LLaMA 7B, the entropy prediction for LLaMA 65B, and the minimum entropy across repetitions from the ICLT method. Minimum entropy is a natural choice based on our intuition behind the ICLT method, and in Appendix \ref{appdx:repetition_context} we include further discussion on the choice of metric. We vary both large and small model sizes and report results in Table \ref{tab:repetition_data}, where we compare the ICLT performance with the SME baseline. 

The intuition behind inserting a separator token after the context, before we re-prompt the model comes from the idea that the next token contains epistemic uncertainty, the relevant information should be present in multiple documents in the training data. As a result, if we were to train the small model longer, such uncertainty could be reduced by learning the relevant information. By inserting a document separator, we are simulating the case when an additional document is present the context, providing the relevant information. In contrast, if the token contains aleatoric-like uncertainty, we should expect that the true language distribution also contains large uncertainty and information provided from a distinct document, the small model will not in-context learn any information. In Table \ref{tab:reptition_separator} we examine the impact of the use of document separator and observe that the use of a separator token is crucial in the performance of the ICLT method. Given the importance of separator token in the ICLT method, we include a preliminary discussion in Appendix \ref{appdx:pythia_analysis} on why it might related to the failure case on Pythia.

\begin{table}[!t]
\caption{\textbf{ICLT results: LLaMA.} ``Band'' denotes lower and upper bounds of small model predictive entropy using LLaMA models of various sizes.}
\label{tab:repetition_data}
\vskip 0.15in
\begin{center}
\begin{small}
\begin{sc}
\begin{tabular}{ccccc|cc}
\toprule
     S & L & Band  &  \multicolumn{2}{c|}{Baseline} &  \multicolumn{2}{c}{Repetition} \\
    \cmidrule(lr){4-7}
      &  &  &  AUC & Acc  & AUC & Acc  \\
    \midrule
     7B & 30B & $[2.0, 3.0)$ &  $0.56$ & $0.55$ &  $0.71$ & $0.67$ \\
     7B & 30B & $[3.0, 4.0)$ &  $0.52$ & $0.55$  & $0.71$ &  $0.66$  \\
     7B & 65B & $[2.0, 3.0)$ &  $0.56$ & $0.55$ &  $0.68$ & $0.66$ \\
     7B & 65B & $[3.0, 4.0)$ &   $0.54$ & $0.54$ &  $0.70$ & $0.68$\\
     30B & 65B & $[2.0, 3.0)$ &  $0.55$ & $0.54$ &  $0.61$ & $0.60$ \\
\bottomrule
\end{tabular}
\end{sc}
\end{small}
\end{center}
\vskip -0.1in
\end{table}

\begin{table*}[!ht]
\caption{\textbf{ICLT results: Ablation on token separator} Ablation study on different types of separators used in between the context and the prompt. The separator is important in the performance of ICLT method. Small model: LLaMA 7B, Large model: LLaMA 30B, entropy band: [2.0, 3.0)]}
\vskip 0.15in
\begin{center}
\begin{small}
\begin{sc}
\begin{tabular}{ccc}
\toprule
Separator Used Type & AUC & Acc \\
\midrule
Original ICLT (BOS) & 0.68 & 63.4 \\
BOS and EOS  & 0.67 &64.0 \\ 
EOS only & 0.62&  57.5\\
None   & 0.56 &52.4 \\
\bottomrule
\end{tabular}
\end{sc}
\end{small}
\end{center}
\label{tab:reptition_separator}
\vskip -0.1in
\end{table*}

\begin{table}[!t]
\caption{\textbf{ICLT results: Pythia (failure case).} ``Band'' denotes lower and upper bounds of small model predictive entropy using Pythia models of various sizes. We use the entropy band $[2.0, 3.0)$.}
\label{tab:repetition_data_pythia}
\vskip 0.15in
\begin{center}
\begin{small}
\begin{sc}
\begin{tabular}{ccccc|cc}
\toprule
     S & L  & Count&  \multicolumn{2}{c|}{Baseline} &  \multicolumn{2}{c}{Repetition} \\
    \cmidrule(lr){4-7}
     &  &   & AUC & Acc & AUC & Acc  \\
    \midrule
     70M & 12B  & 1025 & 0.60 & 0.52 & 0.54 & 0.52  \\
     410M & 12B & 1927 & 0.59 & 0.52 & 0.53 & 0.52  \\
     1B &  12B  & 2314  & 0.54 & 0.54 & 0.54 & 0.54  \\
\bottomrule
\end{tabular}
\end{sc}
\end{small}
\end{center}
\end{table}

\section{Conclusion}


In this paper, we demonstrate that 1) across a variety of text domains and model sizes, LLM embeddings contain enough information to ascertain the certainty of more capable models and 2) this information is sometimes correlated with how willing the model is to copy information in its prompt, permitting unsupervised prediction. In this sense, at least, LLMs ``know what is \textit{knowable},'' distinguishing between prompts for which there is effectively just one correct answer and prompts for which there are many.

Our work is preliminary, and problems remain to be solved before these techniques can be incorporated into practical systems to \textit{e.g.} reduce hallucinations. Some are straightforward engineering challenges; future work is needed to assess performance on more model pairings and datasets,\footnote{In particular, we are interested in measuring how performance evolves as the larger model is increased in size.} tune the number of ``entropy bands'' and thresholds, understand qualitatively where our classifiers fail, and improve classifier performance (both precision and recall) on heavily unbalanced real-world token datasets. Others require more conceptual work. For our initial assumptions to hold, the large models in our pairings should exhibit as little epistemic uncertainty as possible. Whether this goal simply requires increasing the scale of the large model and choosing the right dataset or can be accomplished by other means---perhaps by applying our unsupervised approach to the larger model before using it to generate labels for supervised probes---is currently unclear.

\FloatBarrier
\textbf{Acknowledgements}

We thank Sham Kakade and Garrett Tanzer for useful discussions and feedback on the manuscript.

GA is supported by a fellowship from the Kempner Institute for
the Study of Natural and Artificial Intelligence at Harvard University. NV acknowledges funding from NSF grant DMS-2134157
and DOE grant DE-SC0022199. BB is supported by a Simons Investigator Fellowship, NSF grant DMS-2134157, DARPA grant W911NF2010021, and DOE grant DE-SC0022199. BB is currently affiliated with OpenAI, but this work was done at Harvard. BE acknowledges funding from the NSF Graduate Research Fellowship Program under award DGE-214074, the ONR under award N00014-22-1-2377, and the NSF under award IIS 2229881. This work has been made possible in part by a gift from the Chan Zuckerberg Initiative Foundation to establish the Kempner Institute for the Study of Natural and Artificial Intelligence.

\bibliography{refs}
\bibliographystyle{bibstyle}

\newpage
\appendix
\addcontentsline{toc}{section}{Appendix} 
\part{Appendix} 
\parttoc 

\appendixsection{FAQ}
\label{faq}

In this section, we answer some common questions about our approach.

\textbf{What can be expected to happen as you increase the size of the ``large'' model?
}

A natural question to ask is how the difficulty of the uncertainty classification task is affected by the size of the ``large'' model. In the paper, we provide results using Pythia 12B as the ``large'' model as well as results for the LLaMA pairings 7B/30B and 7B/65B. For both the supervised and unsupervised experiments, there does not appear to be a substantial difference between the two sets of LLaMA figures, either in or out of domain. Supervised results are weaker for the Pythia heads, albeit that those also rely on a smaller ``small'' model (Pythia 1.4B) than the LLaMA heads. While this might simply imply that the 30B LLaMA model is already sufficiently ``knowledgeable'' for our purposes, there are also countervailing reasons to believe that the prediction task grows both easier and more difficult along with the size of the large model.

On the one hand, a larger and better-trained model knows more, and can be expected to have less epistemic uncertainty over the same dataset than a smaller and less capable model. As a result, one can expect to find less noise in the label sets generated by the large model; specifically, there will be fewer tokens in the ``high entropy'' category that actually correspond to ``knowable'' facts. We expect this to simplify the task.

On the other hand, what constitutes a low-entropy ``fact'' for a model may become more subtle as it grows; larger models are better at noticing patterns in text that might at first glance appear ``aleatoric'' (\textit{e.g.} obscure idioms, archaic verb conjugations, and other things we as human readers might not even be aware of). Depending on the makeup of their training sets, larger models can also be expected to have memorized more text, which is potentially another source of noise in the other direction. A memorized ``aleatoric'' passage cannot be distinguished from an instance of epistemic uncertainty using our method, and it is naturally impossible for the small model to predict whether any given passage was memorized by the larger model.

It is worth noting too that the predictions of smaller ``large'' models are closer to those of their respective ``small'' models, improving the quality of trivial baselines based on the small model's prediction. In the limit, where the small model and the large model are the same, it is trivially possible to predict the entropy of the ``large'' model with perfect accuracy.

We are interested in testing our methods with even larger and more knowledgeable ``large'' models (at the scale of \textit{e.g.} ChatGPT) to pin down the net effects.

\textbf{Why would models trained for next token prediction have internal representations of uncertainty?}

Recent work in mechanistic interpretability has identified a number of useful internal representations in large language models, like a ``truthfulness'' direction in activation space \cite{ccs, iti}. In some cases, the intuitive benefits of these mechanisms for next token prediction is clear; internet text, for example, is more often than not correlated with true facts about the world, and so it is in principle helpful for the model to understand the difference between facts and falsehoods. Why it could be useful to form internal representations of different forms of uncertainty is less clear.

One simple explanation involves the fact our unsupervised method depends on: tokens that are more or less ``epistemic'' might correspondingly rely more or less on information earlier on in the prompt. Of course, the entropy of tokens in natural language is not bimodal like that of answer tokens in our synthetic setting. Nevertheless, strictly ``factual'' tokens do exist, and some of the same logic could apply in a ``soft'' way for more ambiguous real tokens. 

\textbf{Is it possible to generate labels without access to a ``large'' model?}

\begin{figure}[!t]
\vskip 0.2in
\begin{center}
\centerline{\includegraphics[width=0.75\columnwidth]{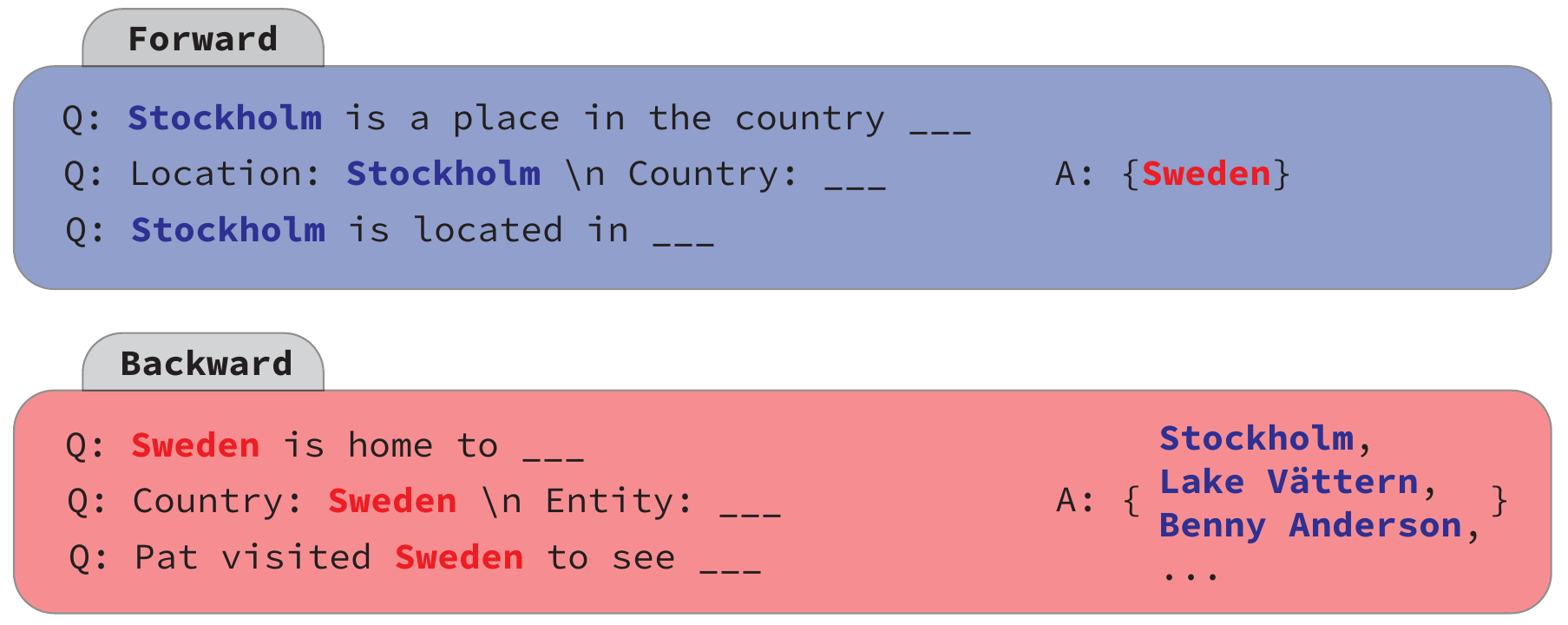}}
\caption{Examples of prompts for the ``country'' Wikidata category. ``Forward'' prompts have one correct answer; ``backward'' prompts have many.}
\label{fig:sample_wikidata}
\end{center}
\vskip -0.2in
\end{figure}

In an attempt to permit evaluation with gold-standard labels for aleatoric and epistemic uncertainty, rather than labels based on discrepancies between small and large language models, we experimented with dataset of synthetic prompts using Wikidata, a knowledge graph derived from Wikipedia. For several hand-picked ``many-to-one'' data categories, containing e.g. book/author or entity/location pairs, we manually wrote a small number of few-shot template prompts \textit{in both directions}. In the ``forward'' case, the model is asked to provide the answer to a question for which it is expected that there is exactly one correct answer (\textit{e.g.} identifying the author of a book). In the ``backward'' case, the prompts are flipped, and the right answer is one of many (\textit{e.g.} identifying a book written by an author). We filtered the pairs such that each ``backward'' prompt has at least five answers in Wikidata. ``Forward'' prompts are intended to elicit epistemic uncertainty, and ``backward'' aleatoric. See Figure \ref{fig:sample_wikidata} for examples of both. 

We did not succeed at producing a diverse enough set of prompts and had issues getting the LLMs to behave as expected on the ones we did write before publication, but we still believe such a dataset could be a useful contribution. We expect that more careful filtering (\textit{e.g.} removing prompts for which a particular ``large'' LLM yields the wrong answer), more specific subcategorization (Wikidata categories tend to be extremely broad, to the point where it is difficult to write natural prompts in both directions), better prompt construction (we tried some simple few-shot settings, but not exhaustively), and potentially leveraging LLMs to produce a larger and more diverse set of prompts would yield improved results. 

\textbf{If internal representations of uncertainty exist, shouldn't RLHF'd models  already have detected and taken advantage of them?}

If levels of epistemic or aleatoric uncertainty can be determined with probes of a model's activations, one might expect that standard techniques for aligning models with human preferences like truthfulness \cite{rlhf_ziegler, rlhf_ouyang, dpo} could pick up on them already, without input from explicit uncertainty classifiers. While this could be the case, these techniques are limited by the ability of human supervisors to identify specific errors, whereas we would ideally want to encourage the models to update their behavior whenever they themselves are uncertain, even in cases where the supervisor couldn't evaluate the correctness of the model's response. From this perspective, uncertainty classifiers may be useful primitives for ``scalable oversight'' \cite{scalable_oversight}, the task of supervising systems that substantially  outperform humans at the particular set of skills being evaluated.

\textbf{Why do baseline results differ from model pairing to model pairing?}

The dataset construction process depends on the entropies of both the small model (via entropy banding) and the large model (via label balancing). As such, the makeup of each test set differs slightly from experiment to experiment.

\textbf{Do the two classes in the binary classification experiments really correspond to aleatoric and epistemic uncertainty?}

The second (higher) bin in our classification experiments is really a catch-all bin for tokens with strictly aleatoric uncertainty, tokens with epistemic uncertainty \textit{and} aleatoric uncertainty (on the part of both models), and tokens where both the large and the small model have significant amounts of epistemic uncertainty. This is one of the reasons why it is important to use a high-quality large model.

\appendixsection{Related work (extended)}
\subsection{Measuring calibration}

Substantial recent work has sought to answer whether large language models are well calibrated in various text domains \cite{desai_calibration, jiang_calibration, anthropic_calibration}. Calibrating language models is closely related to but not identical to the task of distinguishing between different forms of uncertainty, as we have discussed. For example, a model can achieve perfect calibration on balanced multiple choice questions without providing any useful information about its confidence by outputting a uniform distribution over answer choices. Nevertheless, in practice, well-calibrated models can indirectly signal confidence, \textit{e.g.} by predicting confidence scores in text form \cite{anthropic_calibration, tian_just_ask}, and they can be useful for many of the same downstream applications, like selective answering \cite{kamath_selective_answering, varshney_selective_answering, cole_selective_answering}.

\subsection{Combating hallucinations}

While robust predictors of epistemic uncertainty can in principle be used to combat factual hallucinations in language model outputs, methods that do not explicitly rely on uncertainty distinction have also proven effective. Language models can be prompted to be more truthful \cite{si_prompting_gpt}, fine-tuned to output sentences like sentences identified by repeated samples of the model itself to be truthful \cite{tian_finetuning_accuracy}, and modified at inference time to output distributions more consistent with internal truth representations \cite{ccs, iti}. Given that our method tends to detect ``correctable'' tokens---tokens where a small language model is significantly less confident than a larger language model---in imbalanced data with high precision but low recall, many of these approaches could be complementary to ours, regardless of the underlying mechanism by which our method achieves high accuracy.

\cite{ccs} proposes Contrast-Consistent Search (CCS), an unsupervised method for identifying directions in the activation space of an LLM that best separates contradictory answers to questions. These can then be used to elicit latent knowledge from the model, serving as an internal ``truth'' representation analogous to the ``uncertainty'' representation we seek in this work. In follow-up work, \cite{huangdoes} identifies and filters out low-confidence CCS predictions.

\appendixsection{Figure \ref{disagreements} with auto-regressive generation}
\label{disagreements_autoregressive}

\begin{figure*}[!t]
\vskip 0.2in
\begin{center}
\includegraphics[width=0.95\textwidth]{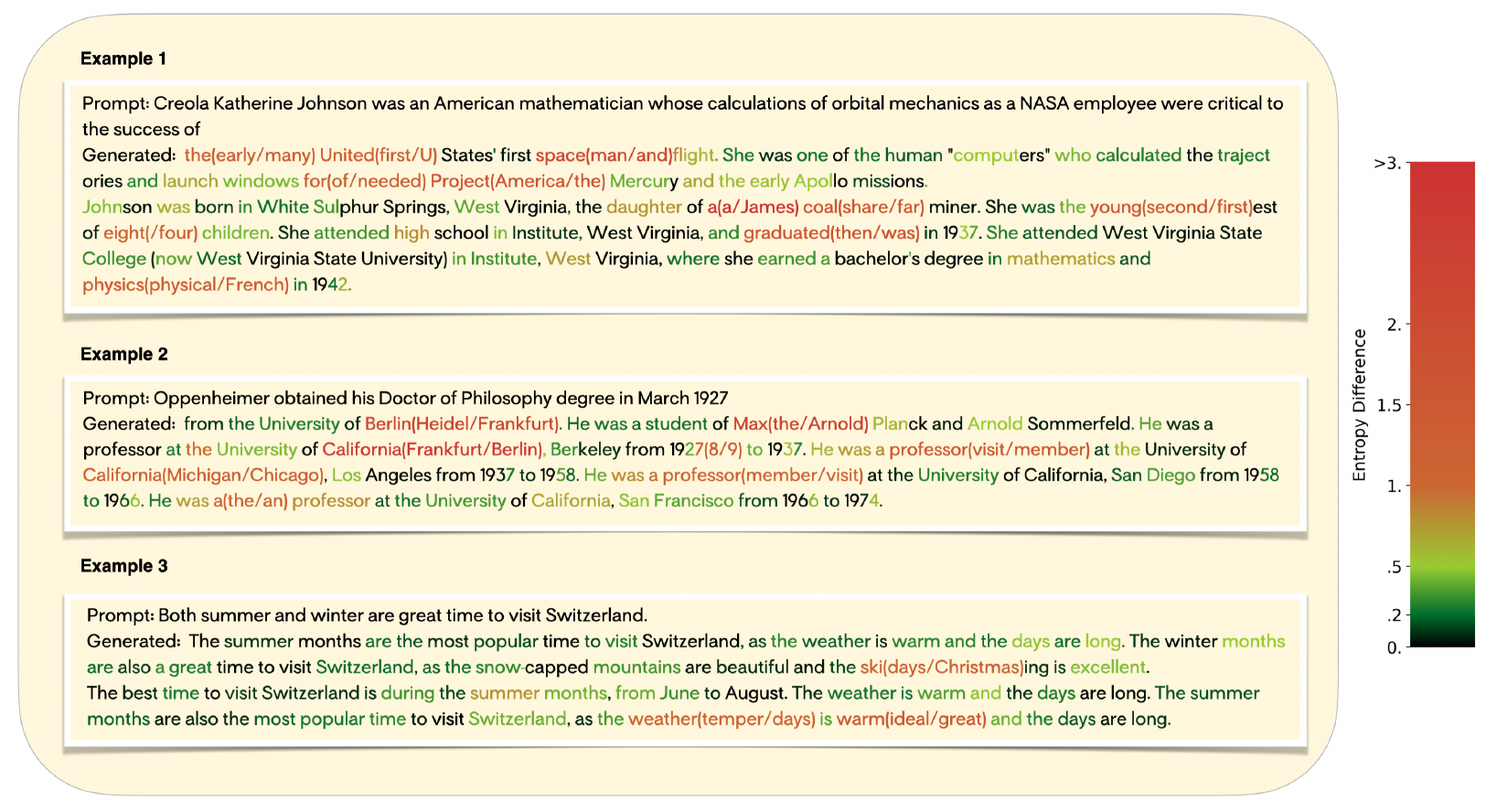}

\caption{A snippet from a Wikipedia article with tokens highlighted wherever the conditional predictive entropies of a pair of small and large language models (LLaMAs 7B and 30B) differ. We show the small model's auto-regressive generation using top 1 token. For any token prediction where the models' entropy difference is larger than 2, we also show small model's next top 2 predictions in parenthesis.}
\label{fig:disagreements_appdx}
\end{center}
\end{figure*}

In Figure \ref{disagreements}, we repeatedly feed snippets from Wikipedia articles into a small (LLaMA 7B) and a large reference model (LLamA 30B). The highlighted text indicates when the two models disagree. Here, we provide an auto-regressive version of the same. In this variant, we use the small model (LLaMA 7B) to autoregressively generate text from given prompts. We then feed the autoregressively generated text from the small model to a large reference model (LLaMA 30B). Figure \ref{fig:disagreements_appdx} shows the text generated by the small model and highlights tokens where two models disagree.

\appendixsection{Training objectives}
\label{training_objectives}

We experimented with the following target values for our classification heads. While each may have unique advantages, we do not find it necessary to stray from large model entropy to achieve high accuracy on our classification tasks.

\begin{description}
    \item \textit{Large model entropy:} The Shannon entropy of $M_{\text{large}}(x)$:
    \begin{equation*}
        H(M_{\text{large}}(x)) = -\sum_{t \in \mathcal{T}} M_{\text{large}}(x)_t \log{M_{\text{large}}(x)_t}
    \end{equation*}
    \item \textit{Log(large model entropy):} The natural log of the Shannon entropy of $M_{\text{large}}(x)$:
    \begin{equation*}
        \log{H(M_{\text{large}}(x))} = \log{\left(-\sum_{t \in \mathcal{T}} M_{\text{large}}(x)_t \log{M_{\text{large}}(x)_t}\right)}
    \end{equation*}
    \item \textit{Jensen-Shannon divergence:} A symmetrized version of the Kullback-Liebler (KL) divergence used to quantify the similarity of two probability distributions over the same support. Letting $D$ represent the KL divergence and $M(x) = 1/2(M_{\text{small}}(x) + M_{\text{large}}(x))$, the Jensen-Shannon divergence is defined as:
    \begin{equation*}
        JSD\infdivx{M_{\text{small}}(x)}{M_{\text{large}}(x)} = 1/2(D\infdivx{M_{\text{small}}(x)}{M(x)} + D\infdivx{M_{\text{small}}(x)}{M(x)})
    \end{equation*}
    \item \textit{Log(Jensen-Shannon divergence):} The natural log of the same.
\end{description}

\appendixsection{Dataset details}

In our experiments, we use a homemade Wikipedia dataset and parts of the Pile evaluation and test sets. We break them down as follows.

Of the 71586 articles in the Wikipedia set (approx. 18.5 million tokens), we set aside 2900 each for validation and testing and use the remaining articles as a training set for our prediction heads. All reported figures use the test set.

From the Pile evaluation set, (approx. 200 million tokens) we sample a validation set of 50k examples, leaving a head training set of 165,000 examples. For parity (and to reduce load on our cluster's filesystem) we subsample the Pile test set by randomly selecting 50k examples. Note that this reduced figure still represents more than 100 million tokens.

\appendixsection{Classification (extended)}
\label{classification_section}

\subsection{Binary classification setup (with gap)}
\label{binary_classification_setup_gap}

By default, we filter tokens for our binary classification task as follows:
\begin{enumerate}
    \item Given some text dataset, run the small and large models on every example and compute the entropy of each model's prediction for each token.
    \item Remove tokens for which the small model's predictive entropy is outside the range $[k, k + 1)$. By default, we set $k = 2$. Note that we bound this value from above because tokens with high small model predictive entropy are more likely to have nonzero large model predictive entropy (see Figure \ref{entropy_is_correlated}). We wish to prevent our heads from depending on this fact.
    \item \label{gap_step} Remove tokens for which the large model entropy is either 1. outside $[0, 0.2)$ or 2. outside some small value $\delta$ of the small model's predictive entropy at that token. We use $\delta = 0.1$. Assign labels to each remaining token corresponding to these two cases.
    \item Independently for each token $t \in \mathcal{V}$, where $\mathcal{V}$ is the models' shared vocabulary, let $T$ be the set of unfiltered tokens for which the previous token in the ground truth is $t$. Balance the two classes within $T$ by discarding random tokens from the larger class (specifically, discard tokens with label $l$ with probability $1 - \min{\{|T_0|, |T_1|\}} / |T_l|$). This procedure is intended to obstruct heads from learning to depend on trivial class-token correlations. 
    
\end{enumerate}

\subsection{Binary classification setup (without gap)}
\label{binary_classification_setup_no_gap}

The setup for the gapless classification task is the same as that described in Section \ref{binary_classification_setup_gap}, minus step \ref{gap_step}.

\subsection{Binary classification full results (gap)}
\label{binary_classification_gap_appendix}

In this section, we include the full set of experiments for the gapped binary classifiers described in Section \ref{classification_with_gap}.

To save space, we use the following shorthands for the names of datasets:
\begin{align*}
    \textsc{Wikipedia} &\rightarrow \textsc{W} \\
    \textsc{Pile} &\rightarrow \textsc{P} \\
    \textsc{Pile (Code)} &\rightarrow \textsc{P (C)} \\
    \textsc{Pile (EuroParl)} &\rightarrow \textsc{P (E)} \\
    \textsc{Pile (Stack Exchange)} &\rightarrow \textsc{P (SE)}
\end{align*}

\begin{figure*}[!t]
\vskip 0.2in
\begin{center}
\centerline{\includegraphics[width=\textwidth]{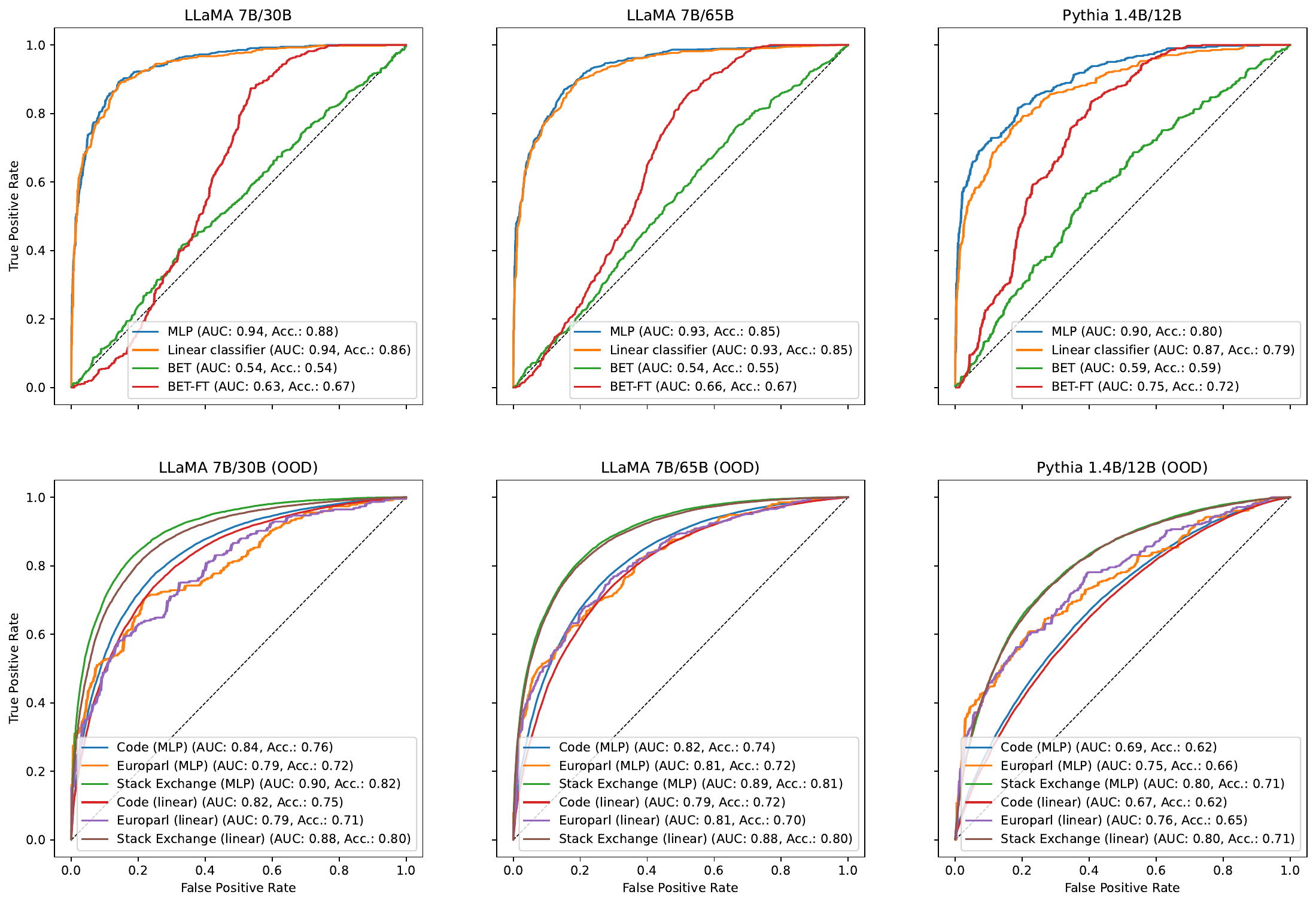}}
\caption{Visualization of the data in Table \ref{tab:binary_classification_gap_results}. \textit{Top:} ROC curves of binary classifiers for large model predictive entropy (with an artificial gap) trained and evaluated with the class-balanced Wikipedia set. Plot titles denote small/large model pairs. Inputs include all tokens for which the small model's predictive entropy is in the range [2, 3). Labels correspond to whether the large model's predictive entropy is near zero or within the same band. Data is provided for three pairs of models (small/large). \textit{Bottom:} The same classifiers evaluated \textit{without additional training} on out-of-distribution data from the Pile.}
\label{fig:binary_classification_gap_curves}
\end{center}
\vskip -0.2in
\end{figure*}

AUROC curves for the experiments described in Table \ref{tab:binary_classification_gap_results} are given in Figure \ref{fig:binary_classification_gap_curves}. Note that the EuroParl set is much smaller than both other two and exhibits much greater variance from classifier to classifier (as seen in \textit{e.g.} the 7B/30B and 7B/65B panels). Also note that the choice of activations used for the Pythia OOD classifiers is suboptimal---see Table \ref{tab:binary_classification_gap_results_appendix_pythia} for a comparison to other embeddings, which achieve much better performance on e.g. the Pile code set.

See Tables \ref{tab:binary_classification_gap_results_appendix_llama_30B}, \ref{tab:binary_classification_gap_results_appendix_llama_65B}, \ref{tab:binary_classification_gap_results_appendix_pythia}, and \ref{tab:binary_classification_gap_results_appendix_llama_2} for LLaMA, Pythia, and Llama 2 results, respectively.

\begin{table*}[!tb]
\caption{LLaMA 7B/30B binary classification results (with a gap) on various test sets. For training details see Section \ref{training_details}, and for baseline definitions, see Section \ref{baselines}. ``Layer'' denotes the layer of the small model from which classifier inputs are drawn.}
\vskip 0.15in
\begin{center}
\begin{small}
\begin{sc}
\begin{tabular}{ccccccc|cc}
\toprule
Model & S & L & Band & Type & Dataset (Train $\rightarrow$ Eval) & Layer & AUC & Acc \\
\midrule
LLaMA & 7B & 30B & $[2, 3)$ & MLP & W $\rightarrow$ W & -1 & 0.92 & 0.84 \\
LLaMA & 7B & 30B & $[2, 3)$ & Linear & W $\rightarrow$ W & -1 & 0.90 & 0.83 \\
LLaMA & 7B & 30B & $[2, 3)$ & MLP & W $\rightarrow$ W & 16 & 0.94 & 0.87 \\
LLaMA & 7B & 30B & $[2, 3)$ & Linear & W $\rightarrow$ W & 16 & 0.94 & 0.86 \\
LLaMA & 7B & 30B & $[2, 3)$ & BET & W $\rightarrow$ W & N/A & 0.54 & 0.54 \\
LLaMA & 7B & 30B & $[2, 3)$ & BET-FT & W $\rightarrow$ W & N/A & 0.63 & 0.67 \\
LLaMA & 7B & 30B & $[2, 3)$ & MLP & P $\rightarrow$ P & -1 & 0.94 & 0.86 \\
LLaMA & 7B & 30B & $[2, 3)$ & Linear & P $\rightarrow$ P & -1 & 0.92 & 0.85 \\
LLaMA & 7B & 30B & $[2, 3)$ & BET & P $\rightarrow$ P & N/A & 0.54 & 0.53 \\
LLaMA & 7B & 30B & $[2, 3)$ & MLP & W $\rightarrow$ P (C) & -1 & 0.79 & 0.71 \\
LLaMA & 7B & 30B & $[2, 3)$ & Linear & W $\rightarrow$ P (C) & -1 & 0.76 & 0.69 \\
LLaMA & 7B & 30B & $[2, 3)$ & MLP & W $\rightarrow$ P (C) & 16 & 0.84 & 0.76 \\
LLaMA & 7B & 30B & $[2, 3)$ & Linear & W $\rightarrow$ P (C) & 16 & 0.82 & 0.75 \\
LLaMA & 7B & 30B & $[2, 3)$ & BET & W $\rightarrow$ P (C) & N/A & 0.52 & 0.53 \\
LLaMA & 7B & 30B & $[2, 3)$ & MLP & W $\rightarrow$ P (E) & -1 & 0.73 & 0.68 \\
LLaMA & 7B & 30B & $[2, 3)$ & Linear & W $\rightarrow$ P (E) & -1 & 0.65 & 0.59 \\
LLaMA & 7B & 30B & $[2, 3)$ & MLP & W $\rightarrow$ P (E) & 16 & 0.79 & 0.70 \\
LLaMA & 7B & 30B & $[2, 3)$ & Linear & W $\rightarrow$ P (E) & 16 & 0.79 & 0.71 \\
LLaMA & 7B & 30B & $[2, 3)$ & BET & W $\rightarrow$ P (E) & N/A & 0.61 & 0.59 \\
LLaMA & 7B & 30B & $[2, 3)$ & MLP & W $\rightarrow$ P (SE) & -1 & 0.88 & 0.79 \\
LLaMA & 7B & 30B & $[2, 3)$ & Linear & W $\rightarrow$ P (SE) & -1 & 0.83 & 0.75 \\
LLaMA & 7B & 30B & $[2, 3)$ & MLP & W $\rightarrow$ P (SE) & 16 & 0.90 & 0.82 \\
LLaMA & 7B & 30B & $[2, 3)$ & Linear & W $\rightarrow$ P (SE) & 16 & 0.88 & 0.80 \\
LLaMA & 7B & 30B & $[2, 3)$ & BET & W $\rightarrow$ P (SE) & N/A & 0.54 & 0.54 \\
\bottomrule
\end{tabular}
\end{sc}
\end{small}
\end{center}
\label{tab:binary_classification_gap_results_appendix_llama_30B}
\vskip -0.1in
\end{table*}

\begin{table*}[!tb]
\caption{LLaMA 7B/65B binary classification results (with a gap) on various test sets. For training details see Section \ref{training_details}, and for baseline definitions, see Section \ref{baselines}. ``Layer'' denotes the layer of the small model from which classifier inputs are drawn.}
\vskip 0.15in
\begin{center}
\begin{small}
\begin{sc}
\begin{tabular}{ccccccc|cc}
\toprule
Model & S & L & Band & Type & Dataset (Train $\rightarrow$ Eval) & Layer & AUC & Acc \\
\midrule
LLaMA & 7B & 65B & $[2, 3)$ & MLP & W $\rightarrow$ W & -1 & 0.92 & 0.85 \\
LLaMA & 7B & 65B & $[2, 3)$ & Linear & W $\rightarrow$ W & -1 & 0.88 & 0.80 \\
LLaMA & 7B & 65B & $[2, 3)$ & MLP & W $\rightarrow$ W & 16 & 0.93 & 0.86 \\
LLaMA & 7B & 65B & $[2, 3)$ & Linear & W $\rightarrow$ W & 16 & 0.93 & 0.85 \\
LLaMA & 7B & 65B & $[2, 3)$ & BET & W $\rightarrow$ W & N/A & 0.54 & 0.55 \\
LLaMA & 7B & 65B & $[2, 3)$ & BET-FT & W $\rightarrow$ W & N/A & 0.66 & 0.67 \\
LLaMA & 7B & 65B & $[2, 3)$ & PIE & W $\rightarrow$ W & N/A & 0.54 & 0.55 \\
LLaMA & 7B & 65B & $[1, 2)$ & MLP & P $\rightarrow$ P & -1 & 0.93 & 0.85 \\
LLaMA & 7B & 65B & $[1, 2)$ & Linear & P $\rightarrow$ P & -1 & 0.90 & 0.82 \\
LLaMA & 7B & 65B & $[1, 2)$ & BET & P $\rightarrow$ P & N/A & 0.55 & 0.54 \\
LLaMA & 7B & 65B & $[2, 3)$ & MLP & P $\rightarrow$ P & -1 & 0.95 & 0.87 \\
LLaMA & 7B & 65B & $[2, 3)$ & Linear & P $\rightarrow$ P & -1 & 0.93 & 0.85 \\
LLaMA & 7B & 65B & $[2, 3)$ & BET & P $\rightarrow$ P & N/A & 0.52 & 0.52 \\
LLaMA & 7B & 65B & $[3, 4)$ & MLP & P $\rightarrow$ P & -1 & 0.95 & 0.87 \\
LLaMA & 7B & 65B & $[3, 4)$ & Linear & P $\rightarrow$ P & -1 & 0.93 & 0.86 \\
LLaMA & 7B & 65B & $[3, 4)$ & BET & P $\rightarrow$ P & N/A & 0.52 & 0.52 \\
LLaMA & 7B & 65B & $[2, 3)$ & MLP & W $\rightarrow$ P (C) & -1 & 0.78 & 0.69 \\
LLaMA & 7B & 65B & $[2, 3)$ & Linear & W $\rightarrow$ P (C) & -1 & 0.73 & 0.66 \\
LLaMA & 7B & 65B & $[2, 3)$ & MLP & W $\rightarrow$ P (C) & 16 & 0.82 & 0.74 \\
LLaMA & 7B & 65B & $[2, 3)$ & Linear & W $\rightarrow$ P (C) & 16 & 0.79 & 0.72 \\
LLaMA & 7B & 65B & $[2, 3)$ & BET & W $\rightarrow$ P (C) & N/A & 0.51 & 0.51 \\
LLaMA & 7B & 65B & $[2, 3)$ & MLP & W $\rightarrow$ P (E) & -1 & 0.73 & 0.64 \\
LLaMA & 7B & 65B & $[2, 3)$ & Linear & W $\rightarrow$ P (E) & -1 & 0.69 & 0.62 \\
LLaMA & 7B & 65B & $[2, 3)$ & MLP & W $\rightarrow$ P (E) & 16 & 0.81 & 0.71 \\
LLaMA & 7B & 65B & $[2, 3)$ & Linear & W $\rightarrow$ P (E) & 16 & 0.81 & 0.70 \\
LLaMA & 7B & 65B & $[2, 3)$ & BET & W $\rightarrow$ P (E) & N/A & 0.55 & 0.55 \\
LLaMA & 7B & 65B & $[2, 3)$ & MLP & W $\rightarrow$ P (SE) & -1 & 0.86 & 0.78 \\
LLaMA & 7B & 65B & $[2, 3)$ & Linear & W $\rightarrow$ P (SE) & -1 & 0.83 & 0.75 \\
LLaMA & 7B & 65B & $[2, 3)$ & MLP & W $\rightarrow$ P (SE) & 16 & 0.89 & 0.81 \\
LLaMA & 7B & 65B & $[2, 3)$ & Linear & W $\rightarrow$ P (SE) & 16 & 0.88 & 0.80 \\
LLaMA & 7B & 65B & $[2, 3)$ & BET & W $\rightarrow$ P (SE) & N/A & 0.53 & 0.53 \\
\bottomrule
\end{tabular}
\end{sc}
\end{small}
\end{center}
\label{tab:binary_classification_gap_results_appendix_llama_65B}
\vskip -0.1in
\end{table*}

\begin{table*}[!tb]
\caption{Pythia binary classification results (with a gap) on various test sets. For training details see Section \ref{training_details}, and for baseline definitions, see Section \ref{baselines}. ``Layer'' denotes the layer of the small model (out of 16) from which classifier inputs are drawn.}
\vskip 0.15in
\begin{center}
\begin{small}
\begin{sc}
\begin{tabular}{ccccccc|cc}
\toprule
Model & S & L & Band & Type & Dataset (Train $\rightarrow$ Eval) & Layer & AUC & Acc \\
\midrule
Pythia & 1.4B & 12B & $[2, 3)$ & MLP & W $\rightarrow$ W & -1 & 0.91 & 0.82 \\
Pythia & 1.4B & 12B & $[2, 3)$ & Linear & W $\rightarrow$ W & -1 & 0.86 & 0.79 \\
Pythia & 1.4B & 12B & $[2, 3)$ & MLP & W $\rightarrow$ W & 8 & 0.90 & 0.81 \\
Pythia & 1.4B & 12B & $[2, 3)$ & Linear & W $\rightarrow$ W & 8 & 0.87 & 0.79 \\
Pythia & 1.4B & 12B & $[2, 3)$ & BET & W $\rightarrow$ W & N/A & 0.59 & 0.59 \\
Pythia & 1.4B & 12B & $[2, 3)$ & BET-FT & W $\rightarrow$ W & N/A & 0.75 & 0.71 \\
Pythia & 1.4B & 12B & $[2, 3)$ & MLP & P $\rightarrow$ P & -1 & 0.93 & 0.86 \\
Pythia & 1.4B & 12B & $[2, 3)$ & Linear & P $\rightarrow$ P & -1 & 0.92 & 0.84 \\
Pythia & 1.4B & 12B & $[2, 3)$ & BET & P $\rightarrow$ P & N/A 
& 0.56 & 0.54 \\
Pythia & 1.4B & 12B & $[2, 3)$ & MLP & W $\rightarrow$ P (C) & -1 & 0.77 & 0.70 \\
Pythia & 1.4B & 12B & $[2, 3)$ & Linear & W $\rightarrow$ P (C) & -1 & 0.75 & 0.67 \\
Pythia & 1.4B & 12B & $[2, 3)$ & MLP & W $\rightarrow$ P (C) & 8 & 0.69 & 0.62 \\
Pythia & 1.4B & 12B & $[2, 3)$ & Linear & W $\rightarrow$ P (C) & 8 & 0.67 & 0.62 \\
Pythia & 1.4B & 12B & $[2, 3)$ & BET & W $\rightarrow$ P (C) & N/A & 0.55 & 0.54 \\
Pythia & 1.4B & 12B & $[2, 3)$ & MLP & W $\rightarrow$ P (E) & -1 & 0.74 & 0.68 \\
Pythia & 1.4B & 12B & $[2, 3)$ & Linear & W $\rightarrow$ P (E) & -1 & 0.68 & 0.62 \\
Pythia & 1.4B & 12B & $[2, 3)$ & MLP & W $\rightarrow$ P (E) & 8 & 0.75 & 0.66 \\
Pythia & 1.4B & 12B & $[2, 3)$ & Linear & W $\rightarrow$ P (E) & 8 & 0.76 & 0.65 \\
Pythia & 1.4B & 12B & $[2, 3)$ & BET & W $\rightarrow$ P (E) & N/A & 0.60 & 0.59 \\
Pythia & 1.4B & 12B & $[2, 3)$ & MLP & W $\rightarrow$ P (SE) & -1 & 0.84 & 0.76 \\
Pythia & 1.4B & 12B & $[2, 3)$ & Linear & W $\rightarrow$ P (SE) & -1 & 0.80 & 0.68 \\
Pythia & 1.4B & 12B & $[2, 3)$ & MLP & W $\rightarrow$ P (SE) & 8 & 0.80 & 0.71 \\
Pythia & 1.4B & 12B & $[2, 3)$ & Linear & W $\rightarrow$ P (SE) & 8 & 0.80 & 0.71 \\
Pythia & 1.4B & 12B & $[2, 3)$ & BET & W $\rightarrow$ P (SE) & N/A & 0.58 & 0.56 \\
\bottomrule
\end{tabular}
\end{sc}
\end{small}
\end{center}
\label{tab:binary_classification_gap_results_appendix_pythia}
\vskip -0.1in
\end{table*}

\begin{table*}[!tb]
\caption{Llama 2 binary classification results (with a gap). For training details see Section \ref{training_details}, and for baseline definitions, see Section \ref{baselines}. Input activations are drawn from the last layer of the small model.}
\vskip 0.15in
\begin{center}
\begin{small}
\begin{sc}
\begin{tabular}{cccccc|cc}
\toprule
Model & S & L & Band & Type & Dataset (Train $\rightarrow$ Eval) & AUC & Acc \\
\midrule
Llama 2 & 7B & 70B & $[1, 2)$ & MLP & W $\rightarrow$ W & 0.94 & 0.87 \\
Llama 2 & 7B & 70B & $[1, 2)$ & Linear & W $\rightarrow$ W & 0.91 & 0.83 \\
Llama 2 & 7B & 70B & $[1, 2)$ & BET & W $\rightarrow$ W & 0.57 & 0.55 \\
Llama 2 & 7B & 70B & $[2, 3)$ & MLP & W $\rightarrow$ W & 0.93 & 0.84 \\
Llama 2 & 7B & 70B & $[2, 3)$ & Linear & W $\rightarrow$ W & 0.90 & 0.83 \\
Llama 2 & 7B & 70B & $[2, 3)$ & BET & W $\rightarrow$ W & 0.55 & 0.55 \\
Llama 2 & 7B & 70B & $[3, 4)$ & MLP & W $\rightarrow$ W & 0.90 & 0.80 \\
Llama 2 & 7B & 70B & $[3, 4)$ & Linear & W $\rightarrow$ W & 0.87 & 0.79 \\
Llama 2 & 7B & 70B & $[3, 4)$ & BET & W $\rightarrow$ W & 0.54 & 0.54 \\
Llama 2 & 7B (chat) & 70B & $[2, 3)$ & MLP & W $\rightarrow$ W & 0.90 & 0.84 \\
Llama 2 & 7B (chat) & 70B & $[2, 3)$ & Linear & W $\rightarrow$ W & 0.90 & 0.83 \\
Llama 2 & 7B (chat) & 70B & $[2, 3)$ & BET & W $\rightarrow$ W & 0.49 & 0.53 \\
\bottomrule
\end{tabular}
\end{sc}
\end{small}
\end{center}
\label{tab:binary_classification_gap_results_appendix_llama_2}
\vskip -0.1in
\end{table*}

\subsection{Binary classification full results (without gap)}

Results for the experiments described in Section \ref{binary_classification_no_gap} are provided in Table \ref{tab:binary_classification_no_gap_results}. While lower than the results on the gapped task, mostly on account of previously absent points near the decision threshold, the classifiers are still accurate, with AUC scores $> 0.8$ across the board compared to baseline scores $< 0.55$. The choice of threshold has a surprisingly small effect on the performance of the final classifiers.

\begin{table*}[!t]
\caption{Binary classification results (without a gap) on the Wikipedia test set (after balancing). ``Threshold'' denotes the boundary between the two bins (in bits). ``Layer'' denotes the layer of the small model (out of 32) from which classifier inputs are drawn.}
\vskip 0.15in
\begin{center}
\begin{small}
\begin{sc}
\begin{tabular}{ccccccc|cc}
\toprule
Model & S & L & Band & Type & Threshold & Layer & AUC & Acc \\
\midrule
LLaMA & 7B & 65B & $[2, 3)$ & MLP & 1 & -1 & 0.83 & 0.76 \\
LLaMA & 7B & 65B & $[2, 3)$ & Linear & 1 & -1 & 0.80 & 0.74 \\
LLaMA & 7B & 65B & $[2, 3)$ & MLP & 1 & 16 & 0.85 & 0.77 \\
LLaMA & 7B & 65B & $[2, 3)$ & Linear & 1 & 16 & 0.83 & 0.75 \\
LLaMA & 7B & 65B & $[2, 3)$ & BET & 1 & N/A & 0.58 & 0.56 \\
LLaMA & 7B & 65B & $[2, 3)$ & MLP & 0.5 & -1 & 0.84 & 0.77 \\
LLaMA & 7B & 65B & $[2, 3)$ & Linear & 0.5 & -1 & 0.83 & 0.75 \\
LLaMA & 7B & 65B & $[2, 3)$ & BET & 0.5 & N/A & 0.55 & 0.54 \\
LLaMA & 7B & 65B & $[2, 3)$ & MLP & 0.2 & -1 & 0.84 & 0.77 \\
LLaMA & 7B & 65B & $[2, 3)$ & Linear & 0.2 & -1 & 0.82 & 0.76 \\
LLaMA & 7B & 65B & $[2, 3)$ & BET & 0.2 & N/A & 0.54 & 0.54 \\
LLaMA & 7B & 65B & $[3, 4)$ & MLP & 1 & -1 & 0.82 & 0.75 \\
LLaMA & 7B & 65B & $[3, 4)$ & Linear & 1 & -1 & 0.81 & 0.73 \\
LLaMA & 7B & 65B & $[3, 4)$ & BET & 1 & N/A & 0.55 & 0.55 \\
LLaMA & 7B & 65B & $[3, 4)$ & MLP & 0.5 & -1 & 0.81 & 0.73 \\
LLaMA & 7B & 65B & $[3, 4)$ & Linear & 0.5 & -1 & 0.80 & 0.73 \\
LLaMA & 7B & 65B & $[3, 4)$ & BET & 0.5 & N/A & 0.52 & 0.53 \\
LLaMA & 7B & 65B & $[3, 4)$ & MLP & 0.2 & -1 & 0.83 & 0.75 \\
LLaMA & 7B & 65B & $[3, 4)$ & Linear & 0.2 & -1 & 0.80 & 0.73 \\
LLaMA & 7B & 65B & $[3, 4)$ & BET & 0.2 & N/A & 0.52 & 0.52 \\
\bottomrule
\end{tabular}
\end{sc}
\end{small}
\end{center}
\label{tab:binary_classification_no_gap_results}
\vskip -0.1in
\end{table*}

\begin{figure}[!t]
\vskip 0.2in
\begin{center}
\centerline{\includegraphics[width=0.6\columnwidth]{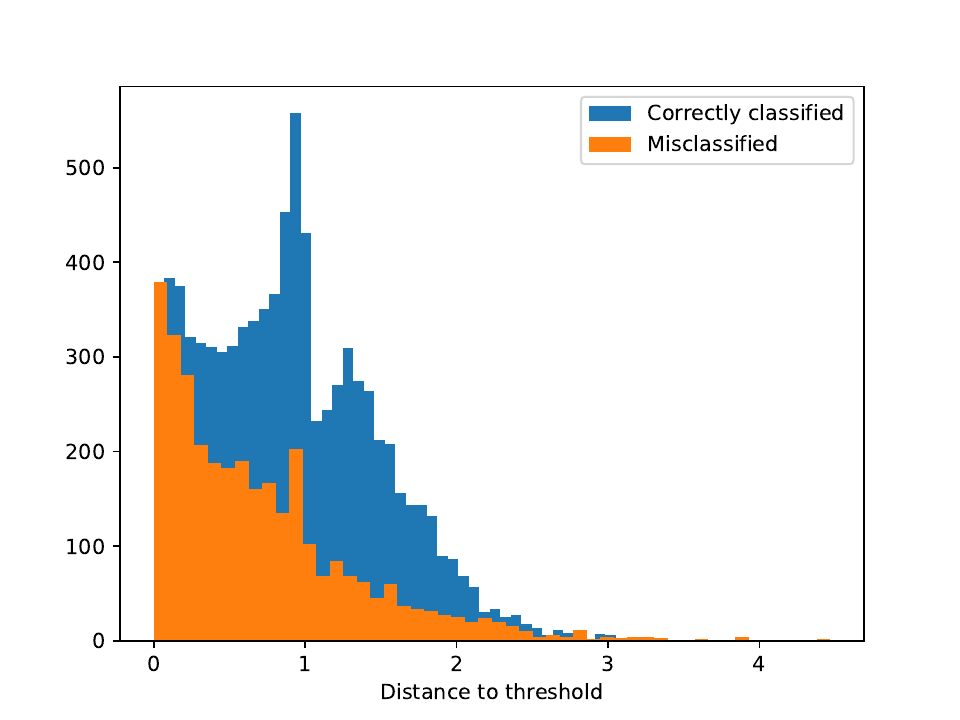}}
\caption{The absolute difference between ground truth large model predictive entropy targets and the classification boundary (1 bit) for examples correctly classified and misclassified by a 7B/65B binary classifier for Wikipedia tokens in the small-model entropy band [2, 3). Near the boundary, accuracy is approximately 50\%, as expected. Accuracy far from the boundary approaches that of classifiers trained with an artificial gap in the distribution of ground truth values.}
\label{fig:classifier_eval_hist}
\end{center}
\vskip -0.2in
\end{figure}

To quantify the performance of these classifiers near their respective decision boundaries, we include a histogram of the distances between target values entropy values and the threshold (in bits) for both misclassified and classified points, using the 7B/65B MLP in the band [2, 3) with a threshold of 1, in Figure \ref{fig:classifier_eval_hist}. As expected, accuracy near the boundary is essentially 50\% and higher farther away from it. Evaluating the same classifier using a filter including a gap with standard hyperparameter settings (i.e. a filter that excludes all points for which the large model's entropy is in the range [0.2, 2) and then rebalances the classes accordingly) yields an AUC score of 0.89 (up from 0.83) and an accuracy of 0.84 (up from 0.76), both comparable to the scores of our ``gapped'' classifiers.

\subsection{Inter-run variance}
\label{inter_run_variance}

\begin{figure}[!t]
\vskip 0.2in
\begin{center}
\centerline{\includegraphics[width=0.7\columnwidth]{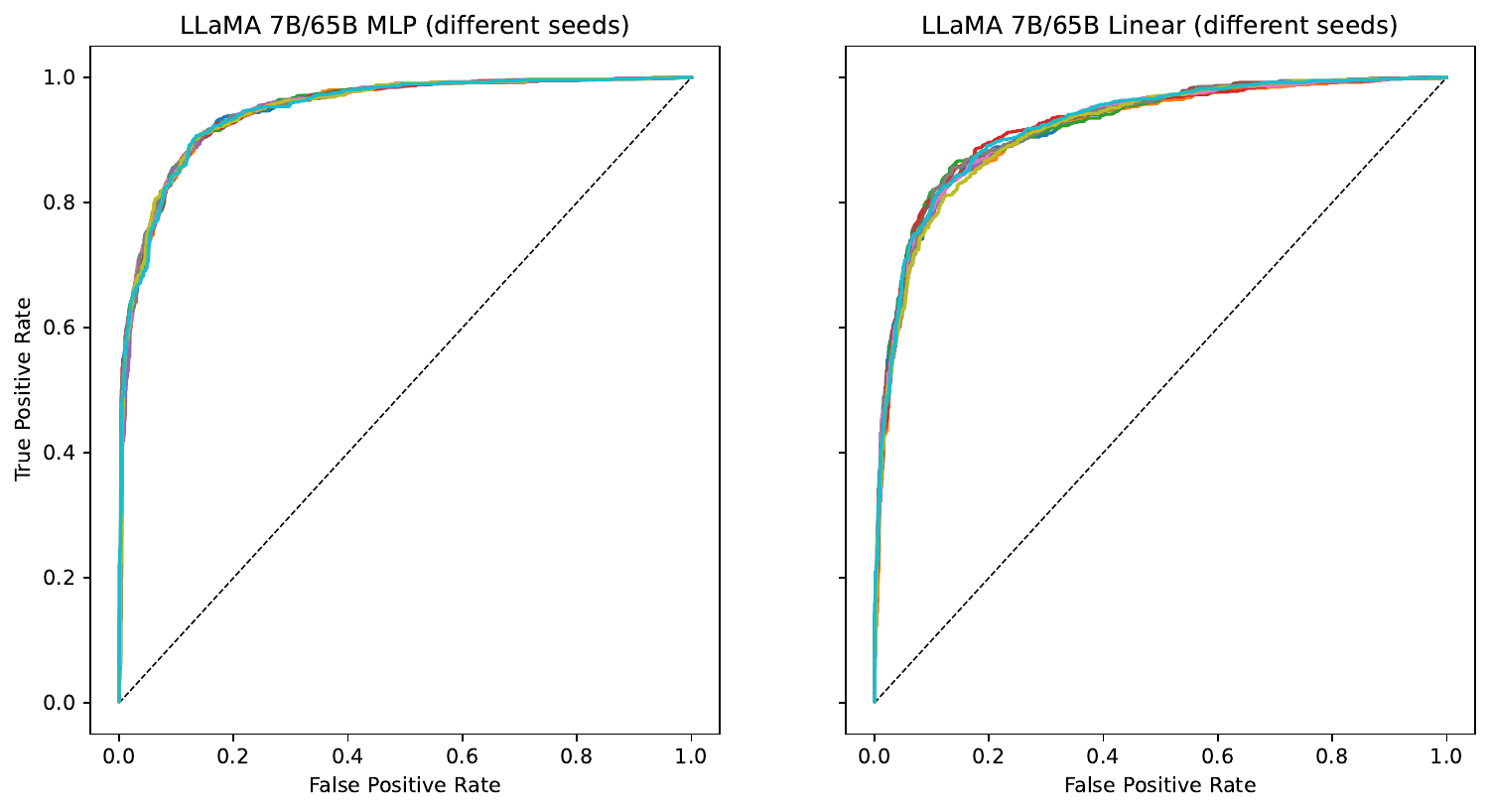}}
\caption{\textbf{Classifier performance is not sensitive to training randomness.} ROC curves for classifiers trained on the Wikipedia set using LLaMA 7B and LLaMA 65B for the binary classification task (with gap). Runs (10 per pane) differ only by the choice of random seed.}
\label{fig:random_seeds}
\end{center}
\vskip -0.2in
\end{figure}

To gauge the sensitivity of the performance of our binary classifiers to training randomness, we train independent LLaMA 7B/65B gapped binary classifiers on the Wikipedia set with different seeds. We include 10 runs each for non-linear and linear classifiers. Seeds determine model initialization and the sequence of training data. For this experiment, dataset filters are held constant. Results are given in Figure \ref{fig:random_seeds}. We observe very little variance between runs; the min and max AUC scores for the MLPs differ by just 0.002.

\subsection{Transfer between models}

How do classifiers trained on one model pairing fare in evaluations with different model pairings? Well.

See Table \ref{tab:binary_classification_gap_results_appendix_unseen_model} for results. We evaluate 7B/30B classifiers using labels generated by the 65B model, and scores are just a few points lower across the board.

\begin{table*}[!tb]
\caption{Binary classification results (with a gap) on an \textit{unseen model pairing}. Classifiers are trained using one large model and evaluated using labels from another. Input activations are drawn from the last layer of the small model.}
\vskip 0.15in
\begin{center}
\begin{small}
\begin{sc}
\begin{tabular}{cccccc|cc}
\toprule
Model & S & L (Train $\rightarrow$ Eval) & Band & Type & Dataset (Train $\rightarrow$ Eval) & AUC & Acc \\
\midrule
LLaMA & 7B & 30B $\rightarrow$ 65B & $[2, 3)$ & MLP & W $\rightarrow$ W & 0.89 & 0.82 \\
LLaMA & 7B & 30B $\rightarrow$ 65B & $[2, 3)$ & Linear & W $\rightarrow$ W & 0.87 & 0.80 \\
LLaMA & 7B & 30B $\rightarrow$ 65B & $[2, 3)$ & MLP & W $\rightarrow$ P & 0.80 & 0.72 \\
LLaMA & 7B & 30B $\rightarrow$ 65B & $[2, 3)$ & Linear & W $\rightarrow$ P & 0.76 & 0.69 \\
LLaMA & 7B & 30B $\rightarrow$ 65B & $[2, 3)$ & MLP & W $\rightarrow$ P (C) & 0.77 & 0.70 \\
LLaMA & 7B & 30B $\rightarrow$ 65B & $[2, 3)$ & Linear & W $\rightarrow$ P (C) & 0.74 & 0.68  \\
LLaMA & 7B & 30B $\rightarrow$ 65B & $[2, 3)$ & MLP & W $\rightarrow$ P (E) & 0.71 & 0.65 \\
LLaMA & 7B & 30B $\rightarrow$ 65B & $[2, 3)$ & Linear & W $\rightarrow$ P (E) & 0.66 & 0.61 \\
LLaMA & 7B & 30B $\rightarrow$ 65B & $[2, 3)$ & MLP & W $\rightarrow$ P (SE) & 0.84 & 0.76 \\
LLaMA & 7B & 30B $\rightarrow$ 65B & $[2, 3)$ & Linear & W $\rightarrow$ P (SE) & 0.82 & 0.74 \\
\bottomrule
\end{tabular}
\end{sc}
\end{small}
\end{center}
\label{tab:binary_classification_gap_results_appendix_unseen_model}
\vskip -0.1in
\end{table*}

\subsection{Choice of embedding}
\label{choice_of_embedding}

\begin{figure}[!t]
\vskip 0.2in
\begin{center}
\centerline{\includegraphics[width=0.49\columnwidth]{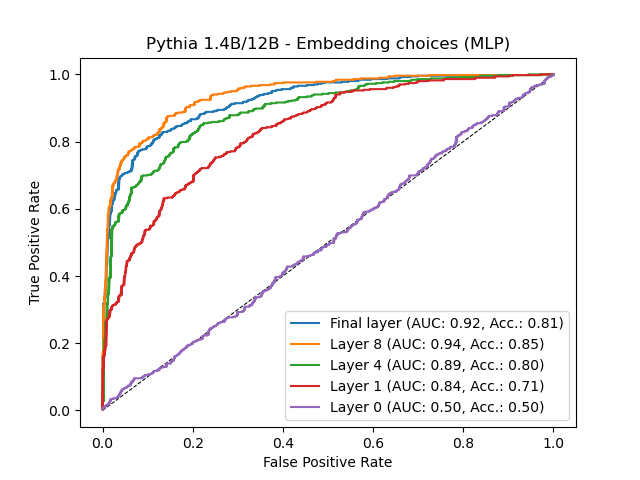}
\includegraphics[width=0.49\columnwidth]{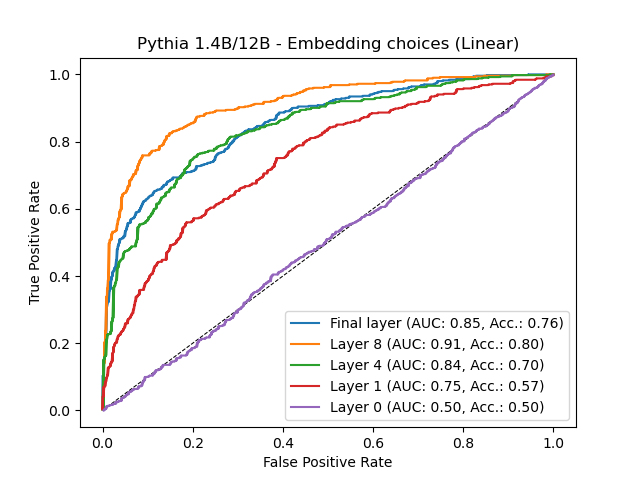}}
\caption{Classification experiments (with a gap) using activations from \textit{different layers} in the small model. \textit{Left:} nonlinear classification. \textit{Right:} linear classification. We use Pythia 1.4B as the small model and Pythia 12B as the large model. Classification heads are trained and tested on Wikipedia data with entropy band $[2.0, 3.0)$.}
\label{supplementary_fig:choice_of_embedding}
\end{center}
\vskip -0.2in
\end{figure}

We train classification heads for the Pythia 1.4B model with embeddings from different layers in each model as inputs. Pythia 1.4B has in total 17 multi-head self-attention layers. In this ablation study, we discovered heads trained on layer 1 (defining layer 17 as the ``final layer" immediately before the logit) already contain enough information on different types of uncertainties in the model. Note that the results from layer 1 can be thought of as a loose ``upper bound'' on the performance of classifiers that learn shallow token-based heuristics as opposed to relying on internal representations in the model; while such representations are not necessarily expected to have formed by the end of the first layer, it is reasonable to expect that the probe can ``see'' all of the tokens in the input by that point. Binary classification results are given in Figure \ref{supplementary_fig:choice_of_embedding} both with linear and non-linear heads. A curious result is that representations from layer 8 seem to outperform the final embeddings we use in the main paper in both cases. Notably, performance out of distribution using embeddings from middle layers tends to be better by an even wider margin (see e.g. Table \ref{tab:binary_classification_gap_results_appendix_llama_65B}). We expect that more exhaustive tuning here would be fruitful.

\subsection{Choice of checkpoint}

We take advantage of the fact that the Pythia models were released along with intermediate checkpoints from training to test the effect of ``rewinding'' the small model on classification performance. Results are given in Table \ref{tab:pythia_back_in_time}. Rewinding back to step 9000 does not have a discernible effect on the quality of the classifiers, perhaps because reducing the quality of the ``small'' model allows more obvious ``epistemic'' tokens to pass the evaluation set filter. However, classifiers were more unstable for the smallest revision, for which the MLP repeatedly failed to converge. Also, classifiers trained on the oldest revisions do not transfer as well to new distributions.

\begin{table*}[!tb]
\caption{Binary classification results (without a gap) on the Wikipedia test set using early checkpoints of the ``small'' model. ``Step'' denotes the Pythia training step of the small model checkpoint. Input activations are drawn from the last layer of the small model.}
\vskip 0.15in
\begin{center}
\begin{small}
\begin{sc}
\begin{tabular}{ccccccc|cc}
\toprule
Model & S & Step & L & Band & Type & Dataset (Train $\rightarrow$ Eval) & AUC & Acc \\
\midrule
Pythia & 1.4B & 9000 & 12B & $[2, 3)$ & MLP & W $\rightarrow$ W & 0.54 & 0.51 \\
Pythia & 1.4B & 9000 & 12B & $[2, 3)$ & Linear & W $\rightarrow$ W & 0.91 & 0.84 \\
Pythia & 1.4B & 9000 & 12B & $[2, 3)$ & BET & W $\rightarrow$ W & 0.46 & 0.66 \\
Pythia & 1.4B & 9000 & 12B & $[2, 3)$ & Linear & W $\rightarrow$ P (C) & 0.57 & 0.56 \\
Pythia & 1.4B & 9000 & 12B & $[2, 3)$ & Linear & W $\rightarrow$ P (E) & 0.68 & 0.64 \\
Pythia & 1.4B & 9000 & 12B & $[2, 3)$ & Linear & W $\rightarrow$ P (SE) & 0.69 & 0.66 \\

Pythia & 1.4B & 18000 & 12B & $[2, 3)$ & MLP & W $\rightarrow$ W & 0.92 & 0.84 \\
Pythia & 1.4B & 18000 & 12B & $[2, 3)$ & Linear & W $\rightarrow$ W & 0.89 & 0.80 \\
Pythia & 1.4B & 18000 & 12B & $[2, 3)$ & BET & W $\rightarrow$ W & 0.46 & 0.50 \\

Pythia & 1.4B & 35000 & 12B & $[2, 3)$ & MLP & W $\rightarrow$ W & 0.92 & 0.81 \\
Pythia & 1.4B & 35000 & 12B & $[2, 3)$ & Linear & W $\rightarrow$ W & 0.86 & 0.76 \\
Pythia & 1.4B & 35000 & 12B & $[2, 3)$ & BET & W $\rightarrow$ W & 0.48 & 0.50 \\

Pythia & 1.4B & 35000 & 12B & $[2, 3)$ & MLP & W $\rightarrow$ W & 0.76 & 0.81 \\
Pythia & 1.4B & 35000 & 12B & $[2, 3)$ & Linear & W $\rightarrow$ W & 0.76 & 0.69 \\
Pythia & 1.4B & 35000 & 12B & $[2, 3)$ & BET & W $\rightarrow$ W & 0.48 & 0.50 \\

Pythia & 1.4B & 70000 & 12B & $[2, 3)$ & MLP & W $\rightarrow$ W & 0.87 & 0.78 \\
Pythia & 1.4B & 70000 & 12B & $[2, 3)$ & Linear & W $\rightarrow$ W & 0.76 & 0.69 \\
Pythia & 1.4B & 70000 & 12B & $[2, 3)$ & BET & W $\rightarrow$ W & 0.54 & 0.53 \\

Pythia & 1.4B & 143000 & 12B & $[2, 3)$ & MLP & W $\rightarrow$ W & 0.91 & 0.82 \\
Pythia & 1.4B & 143000 & 12B & $[2, 3)$ & Linear & W $\rightarrow$ W & 0.86 & 0.79 \\
Pythia & 1.4B & 143000 & 12B & $[2, 3)$ & BET & W $\rightarrow$ W & 0.59 & 0.59 \\
\bottomrule
\end{tabular}
\end{sc}
\end{small}
\end{center}
\label{tab:pythia_back_in_time}
\vskip -0.1in
\end{table*}

\appendixsection{Regressions}
\label{regression_section}

Because thresholds for ``near-zero'' entropy vary from domain to domain, accurate regressions for our tasks would be more immediately useful than simple binary classifiers. In this section, we experiment therewith.

\begin{figure}[!t]
\vskip 0.2in
\begin{center}
\centerline{\includegraphics[width=0.6\columnwidth]{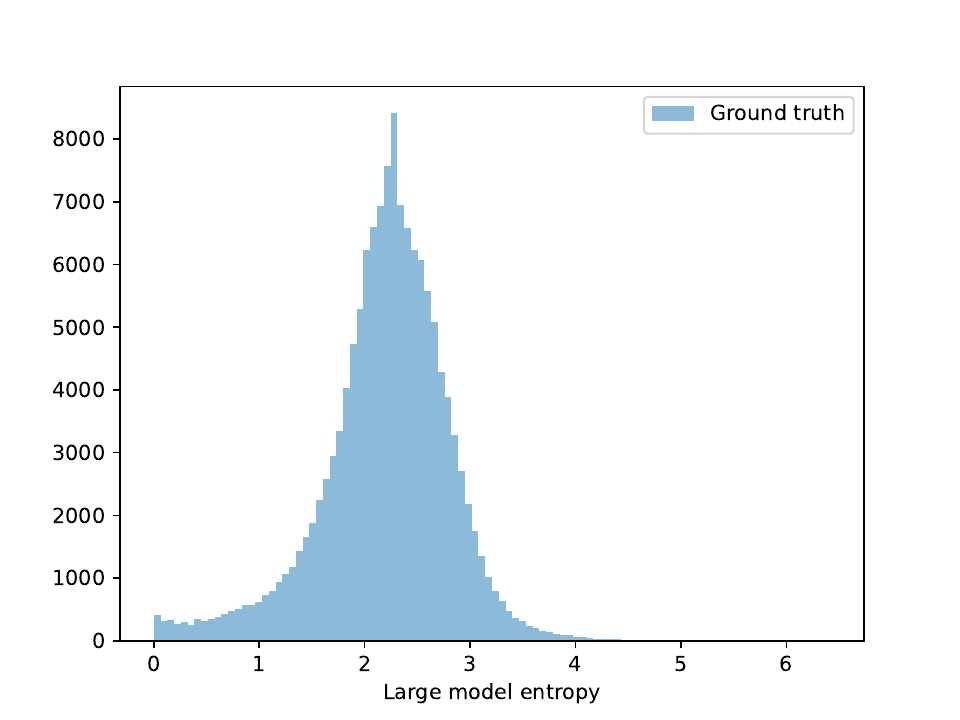}}
\caption{Distribution of target values in the small model entropy 
 band [2,3) of the Wikipedia validation set. Most values are clustered within the small entropy band, and few are significantly smaller.}
\label{fig:regression_val_hist_gt}
\end{center}
\vskip -0.2in
\end{figure}

An immediate challenge is that the distribution of target values is extremely imbalanced (see Figure \ref{fig:regression_val_hist_gt}), even within 1-bit small-model entropy bands, while at the same time we care the most about examples for which the large model has small entropy conditioned on the small model being uncertain (for the same reasons we discuss in the main paper, \textit{e.g.} that we assume the small model is decently well-calibrated and is rarely ``delusionally confident''). We attempt the following interventions to force our regressions to learn something akin to what our classifiers do:

\begin{itemize}
\item \textbf{Upsampling (U):} Somewhat similarly to \cite{imbalanced_regression}, within a small-model entropy band, we upsample the (rare) tokens for which the target value is significantly smaller than the mean. Concretely, we set the probability of accepting some uniformly sampled prompt $x$ to
\begin{equation*}
    P(x) = \min{\left\{1, \frac{1}{\max{\{\epsilon, \alpha * H(L(t))\}}}\right\}}
\end{equation*}
where $H$ is the entropy function, $L$ is the label-generating large model, as before, and $\alpha$ and $\epsilon$ are tunable constants that depend on the degree of imbalance in the training distribution.

\item \textbf{Punishing underestimates (PU):} Of course, upsampling low-frequency tokens with small target values risks simply shifting the distribution of predictions on imbalanced data left, drowning useful signal. To improve ``precision'' for tokens with a small target value, we add a term to the loss to punish underestimates of the target value. For predictions $x$ and target values $y$, we compute the loss as follows:
\begin{equation*}
    \mathcal{L}(x, y) = \underbrace{(x - y)^2}_{\text{squared error}} + \alpha\underbrace{(\max{\{y - x, 0\}})^2}_{\text{squared underestimate}}
\end{equation*}
where $\alpha$ is again a tunable constant.
\end{itemize}

\begin{table*}[!t]
\caption{Regressions trained on Wikipedia data and evaluated on the Wikipedia test set. MSE is standard mean squared error on the test set. SME is the ``small model entropy'' baseline. We evaluate the performance of the regressions as binary classifiers with various thresholds at inference time.}
\vskip 0.15in
\begin{center}
\begin{small}
\begin{sc}
\begin{tabular}{ccccc|c|ccc}
\toprule
Model & S & L & Band & Type & MSE & Threshold & Precision & Recall \\
\midrule
LLaMA & 7B & 65B & $[2, 3)$ & MLP & 0.29 & 1 & 0.70 & 0.14 \\
LLaMA & 7B & 65B & $[2, 3)$ & MLP & 0.29 & 0.5 & 0.61 & 0.08 \\
LLaMA & 7B & 65B & $[2, 3)$ & MLP & 0.29 & 0.2 & 0.46 & 0.06 \\

LLaMA & 7B & 65B & $[2, 3)$ & Linear & 0.37 & 1 & 0.69 & 0.02 \\
LLaMA & 7B & 65B & $[2, 3)$ & Linear & 0.37 & 0.5 & 0.75 & 0.001 \\
LLaMA & 7B & 65B & $[2, 3)$ & Linear & 0.37 & 0.5 & N/A & 0 \\

LLaMA & 7B & 65B & $[2, 3)$ & MLP (+ PU) & 0.33 & 1 & 0.74 & 0.11 \\
LLaMA & 7B & 65B & $[2, 3)$ & MLP (+ PU) & 0.33 & 0.5 & 0.71 & 0.06 \\
LLaMA & 7B & 65B & $[2, 3)$ & MLP (+ PU) & 0.33 & 0.2 & 0.61 & 0.05 \\

LLaMA & 7B & 65B & $[2, 3)$ & Linear (+ PU) & 0.40 & 1 & 0.48 & 0.01 \\
LLaMA & 7B & 65B & $[2, 3)$ & Linear (+ PU) & 0.40 & 0.5 & 0.23 & 0.003 \\
LLaMA & 7B & 65B & $[2, 3)$ & Linear (+ PU) & 0.40 & 0.5 & 0.29 & 0.002 \\

LLaMA & 7B & 65B & $[2, 3)$ & MLP (+ U, PU) & 0.41 & 1 & 0.49 & 0.27 \\
LLaMA & 7B & 65B & $[2, 3)$ & MLP (+ U, PU) & 0.41 & 0.5 & 0.57 & 0.08 \\
LLaMA & 7B & 65B & $[2, 3)$ & MLP (+ U, PU) & 0.41 & 0.2 & 0.65 & 0.03 \\

LLaMA & 7B & 65B & $[2, 3)$ & Linear (+ U, PU) & 0.51 & 1 & 0.38 & 0.26 \\
LLaMA & 7B & 65B & $[2, 3)$ & Linear (+ U, PU) & 0.51 & 0.5 & 0.31 & 0.07 \\
LLaMA & 7B & 65B & $[2, 3)$ & Linear (+ U, PU) & 0.51 & 0.5 & 0.28 & 0.03 \\

LLaMA & 7B & 65B & $[2, 3)$ & SME & 0.39 & - & - & - \\

\bottomrule
\end{tabular}
\end{sc}
\end{small}
\end{center}
\label{tab:regression_results}
\vskip -0.1in
\end{table*}

Results for Wikipedia are given in Table \ref{tab:regression_results}. Clearly, while results are nontrivial, more work is needed to make the regressions usable in the imbalanced case.

\appendixsection{Unsupervised classification (extended)}
\label{unsupervised_appendix}

\subsection{Additional ICLT results}
\begin{figure}[!t]
\vskip 0.2in
\begin{center}
\centerline{\includegraphics[width=0.4\columnwidth]{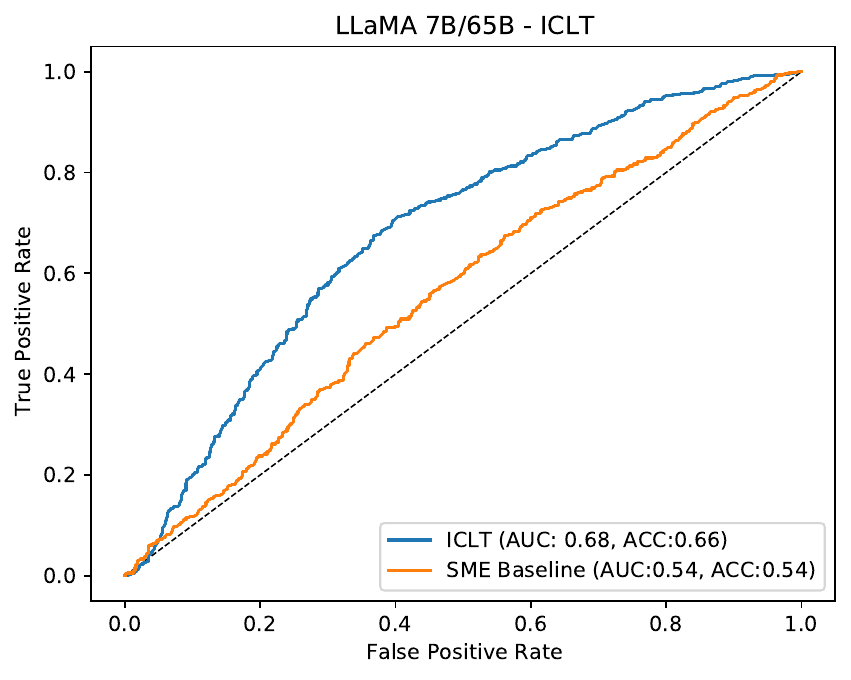}
\includegraphics[width=0.4\columnwidth]{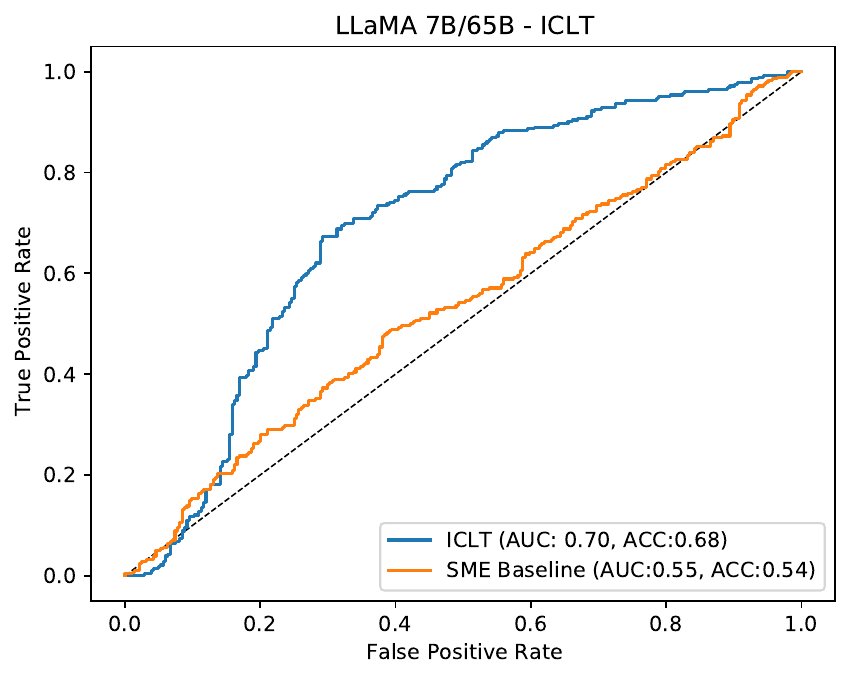}}
\caption{ROC curves for ICLT method compared to SME baseline. \textit{Left:} Entropy bin $[2-3)$. \textit{Right:} Entropy bin: $[3-4)$}
\label{fig:reptition_roc}
\end{center}
\vskip -0.2in
\end{figure}

\begin{figure}[!th]
\vskip 0.2in
\begin{center}
\centerline{
\includegraphics[width=0.8\columnwidth]{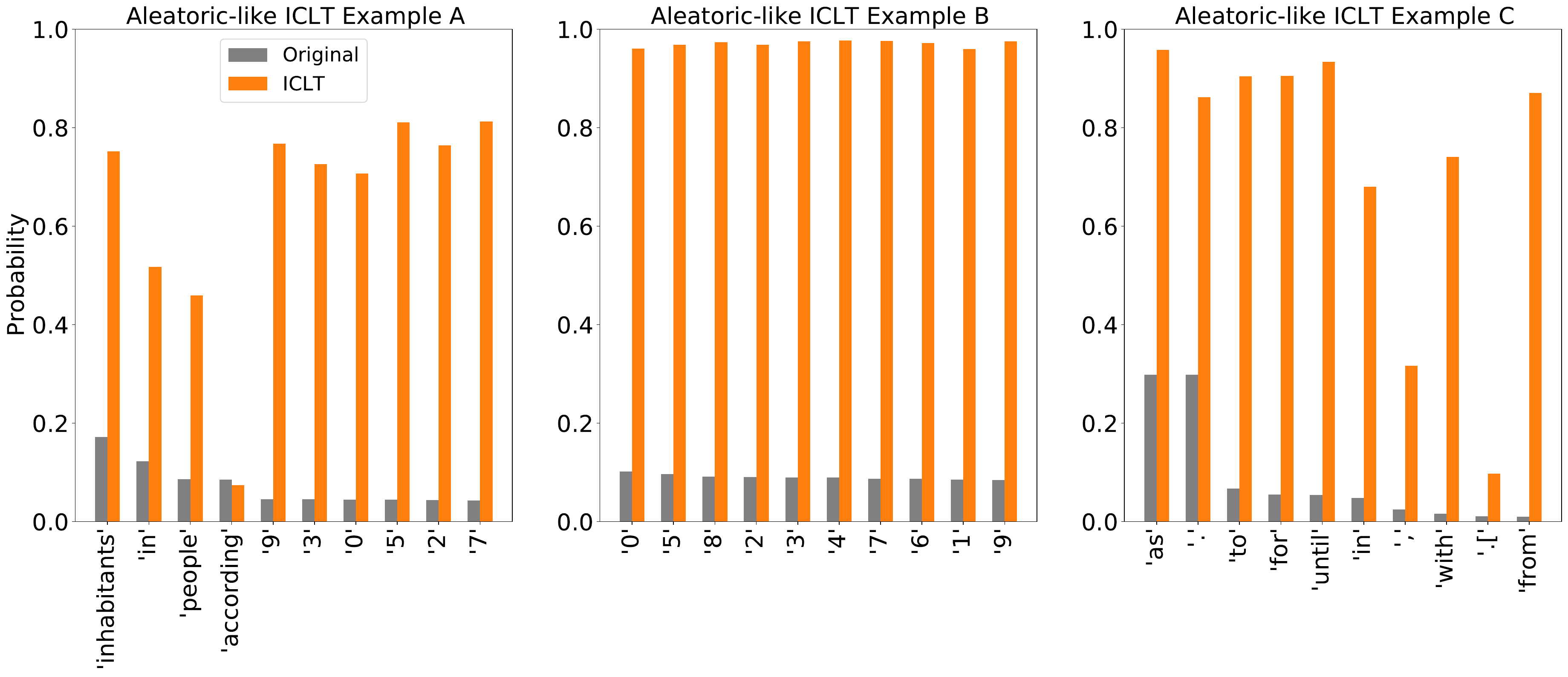}
}
\caption{Additional examples for the ICLT method using LLaMA 7B model to predict LLaMA 30B's entropy. In these examples, LLaMA 30B is unsure about the next token, but the ICLT fails at classifying them as aleatoric.}
\label{fig:repetition_lessgood_aleatoric}
\end{center}

\vskip 0.2in
\begin{center}

\centerline{
\includegraphics[width=0.8\columnwidth]{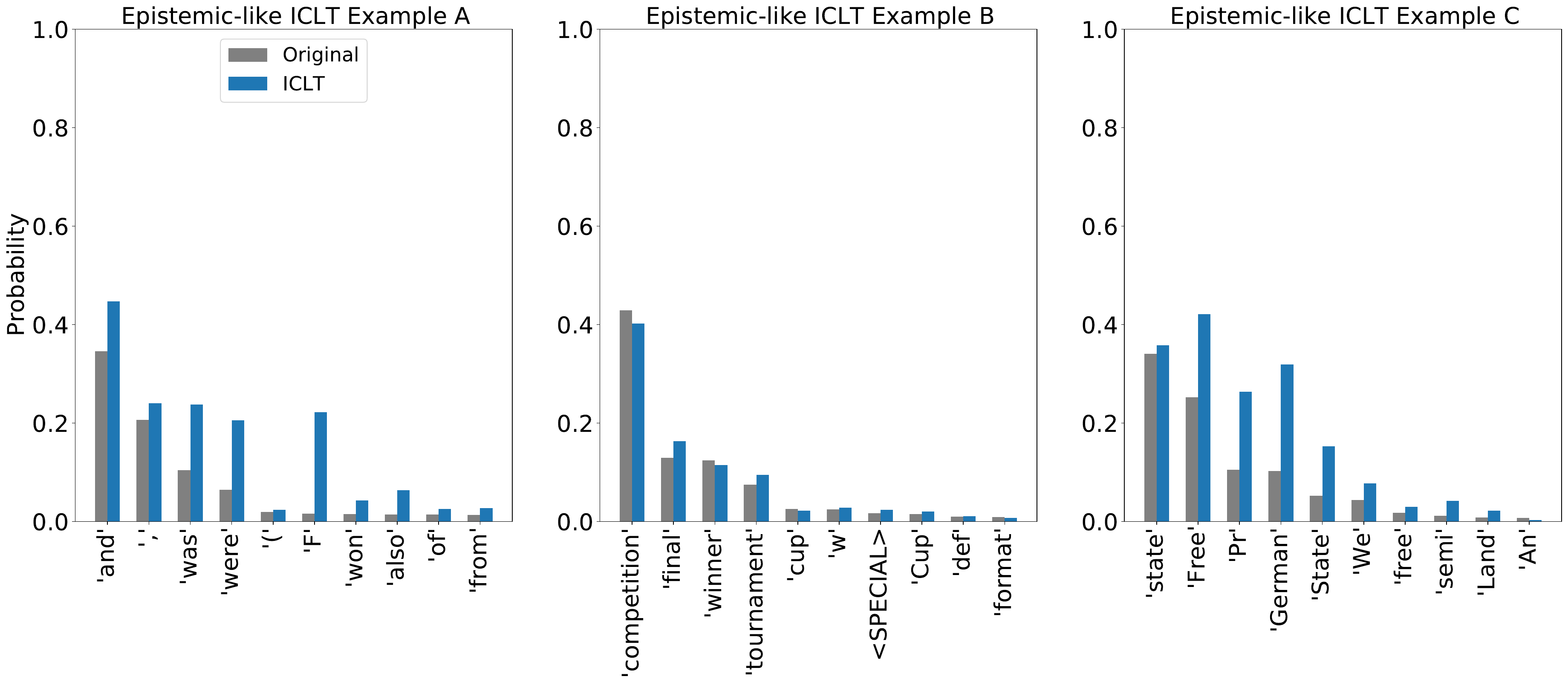}
}
\caption{Additional examples for the ICLT method using LLaMA 7B model to predict LLaMA 30B's entropy. In these examples, LLaMA 30B is sure about the next token, making the uncertainty epistemic, but ICLT fails at classifying examples as such.}
\label{fig:repetition_lessgood_epistemic}
\end{center}
\vskip -0.2in
\end{figure}

\label{appdx:rep_additional_examples}
In Figure \ref{fig:reptition_roc}, we show two sample ROC curves of the unsupervised method with model pair LLaMA 7B/65B under two entropy bins. We can see that the ICLT method consistently outperforms the simple SME baseline. Figure \ref{fig:reptition_synetheic} shows two cases where the ICLT method works as expected (i.e, repeating only when the uncertainty is epistemic). However, those two examples do not paint a complete picture of the ICLT method. In Figure \ref{fig:repetition_lessgood_aleatoric} and Figure \ref{fig:repetition_lessgood_epistemic} we show additional examples where ICLT works less well. Broadly speaking, this often occurs where the large model exhibits epistemic uncertainty.

Below are the prompts used for the six examples given in Figure \ref{fig:repetition_lessgood_aleatoric} and Figure \ref{fig:repetition_lessgood_epistemic}. 

\begin{itemize}
      \item \textbf{Figure \ref{fig:repetition_lessgood_aleatoric}:} Examples where the large model is uncertain about next token prediction (aleatoric)
      \begin{itemize}
          \item \textbf{Example A}  ``Al Mudhafr Al-Asfal is a sub-district located in the Al Bayda District, Al Bayda Governorate, Yemen. Al Mudhafr Al-Asfal had a population of 3527''
         \item \textbf{Example B}``Anastas Ngjela (3 March 1934 – 2''
         \item \textbf{Example C} ``Australian Irish Sign Language or AISL is a minority sign language in Australia. As a Francosign language, it is related to French Sign Language as opposed to Auslan which is a Banzsl language which is related to British Sign Language. AISL was brought to Australia from Ireland in 1875 by a group of Dominican nuns (including a Deaf nun) where three schools were established and used AISL''
    \end{itemize}
    
    \item \textbf{Figure \ref{fig:repetition_lessgood_epistemic}:} Examples where the large model is certain about next token prediction (epistemic)
    \begin{itemize}
    
        \item \textbf{Example A} ``The 1987–88 Gamma Ethniki was the fifth season since the official establishment of the third tier of Greek football in 1983. Atromitos and Makedonikos were crowned champions in `Southern' and `Northern' Group respectively, thus winning promotion to Beta Ethniki. Rethymniak"
         \item \textbf{Example B} ``The 1921–22 City Cup was the 24th edition of the City Cup, a cup competition in Northern Irish football. The"
          \item \textbf{Example C} ``The June 1924 Anhalt state election was held on 22 June 1924 to elect the 36 members of the Landtag of the"
      \end{itemize}
\end{itemize}

\subsection{Ablation study on context provided for ICLT}
\label{appdx:repetition_context}
To further understand the ICLT method, we conducted ablation studies on different contexts provided with full results shown in Table \ref{tab:reptition_context}. We discovered that in general, the ICLT method is not very sensitive to the information provided in the context. In particular, we experimented the following changes to the context provided. 

\textbf{Additional Context} We provided additional context by allowing the small model to auto-repressively generate next token until it outputs an period token or an $\langle \texttt{EOS}\rangle$ token to indicate the end of sentence. Then we prepended the completed sentence as the context information before the original prompt, and feed back into the model again for next-token generation. 

\textbf{Irrelevant Context} We provided irrelevant context after the relevant context and before we repeat the promot: \
\begin{equation}
    \texttt{prompt} + \texttt{relevant context} + \texttt{irrelevant information} +  \texttt{prompt} 
\end{equation}

\textbf{None top $k$} Instead of sampling top $k$ tokens as to generate the context, we used random tokens to generate the context. 

\begin{table*}[!tb]
\caption{ICLT Method ablation on differen kinds of context provided. Overall, the method is not very sentitive to the context provided.}
\vskip 0.15in
\begin{center}
\begin{small}
\begin{sc}
\begin{tabular}{ccc}
\toprule
Ablation Type & AUC & Acc \\
\midrule
Original ICLT & 0.68 & 63.4 \\
Additional context &0.70 & 64.5 \\
Irrelevant context &  0.64& 62.1  \\
Top 1 & 0.65 & 61.4\\ 
Top 5  & 0.67 & 62.9 \\ 
Top 20  & 0.68 & 63.5 \\ 
Random 10 &0.61 & 60.2\\ 
\bottomrule
\end{tabular}
\end{sc}
\end{small}
\end{center}
\label{tab:reptition_context}
\vskip -0.1in
\end{table*}

\textbf{Metric choice} The choice of using minimum entropy as our metric choice comes from the ablation study above. By using the top $k$ generations, one might want to consider using the original softmax probability as a weighting scheme. Some examples might include: weighted entropy from the ICLT generation or mutual information between the original and the ICLT generation. Because the ablation on context information suggests that the ICLT method is not sensitive to the specific kind of context provided, a natural metric choice for the unsupervised task is to exclude any information derived from the context. Furthermore, from the synthetic set-up to examining examples, our understanding is that we are looking for the copying (\textit{i.e.} learning) behaviors from the context as an indicator for epistemic uncertainty. Therefore, we used the minimum entropy among all contexts as a metric. 



\subsection{Why does ICLT fail on Pythia?}

\label{appdx:pythia_analysis}
\begin{figure}[!tb]
\vskip 0.2in
\begin{center}
\centerline{\includegraphics[width=0.4\textwidth]{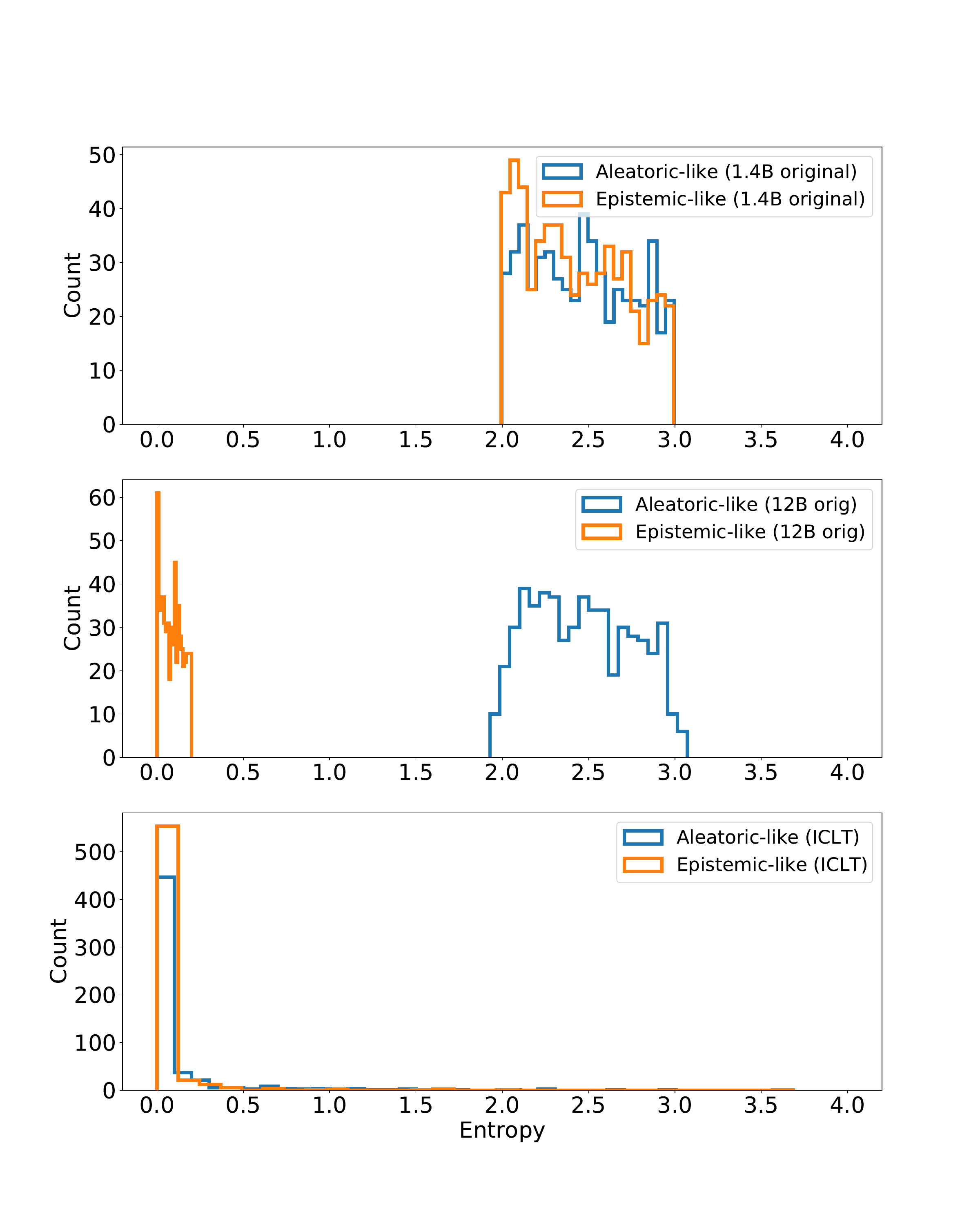}}
\caption{ICLT method on Pythia 1.4B using Pythia 12B to generate labels with Pile test set. \textit{Top:} original entropy prediction by 1.4B model. \textit{Middle:} entropy tagging with 12B model. \textit{Bottom:} Minimum entropy from ICLT method on 1B.}
\label{fig:reptition_entropy_pythia}
\end{center}
\vskip -0.2in
\end{figure}

Pythia has the tendency to repeat everything verbatim from the context regardless of the prompt, as shown in Figure \ref{fig:reptition_entropy_pythia}. Furthermore, such behavior exists both on Wikipedia dataset and Pile test set (shown in Tabel \ref{tab:repetition_pythia_pile}), and exists on all Pythia model sizes. While we did not reach a clear answer on the cause, we discovered that the document separation tokens (i.e., special tokens such as $\langle \texttt{EOS} \rangle$ and $\langle \texttt{BOS} \rangle$ ) play an important role in the success in ICLT method on LLaMA, shown in Table \ref{tab:reptition_separator}. LLaMA uses a byte pair encoding (BPE) tokenizer that has $\langle \texttt{EOS} \rangle$ and $\langle \texttt{BOS} \rangle$ as two separate tokens indicating the beginning and end of the document. On the other hand, Pythia uses a BPE tokenizer \cite{pythia} specifically trained on Pile, and only end of text is indicated.  In our original setup, we use $\langle \texttt{BOS} \rangle$ to indicate the document separation boundary. Inserting additional $\langle \texttt{EOS} \rangle$ token does not affect the performawnce. However, replacing the $\langle \texttt{BOS} \rangle$ with an $\langle \texttt{EOS} \rangle$ token inside the ICLT context significantly affects the model's behavior. Finally, removing any document boundaries (``None" in Table \ref{tab:reptition_separator}) leads to ICLT failing completely. We suspect that the use of different document separators during pre-training has an impact on model's in-context learning behaviors. 

\begin{table}[!tb]
\caption{Unsupervised ICLT method does not work on Pyhtia: using small models (Pythia 70M, 410M, and 1.4B) to classify when large model (Pythia 12B) is confident. Entropy band $[2.0, 3.0)$. Dataset: Pile validation set. Baseline: SME}
\label{tab:repetition_pythia_pile}
\vskip 0.15in
\begin{center}
\begin{small}
\begin{sc}
\begin{tabular}{ccccc|cc}
\toprule
     S & L  & Count&  \multicolumn{2}{c|}{Baseline} &  \multicolumn{2}{c}{Repetition} \\
    \cmidrule(lr){4-7}
     &  &   & Acc & AUC  & Acc & AUC  \\
    \midrule
     70M & 12B  & 1574 & 56.2 & 0.51 & 51.1 &  0.58 \\
     410M & 12B & 1042 &  55.2 & 0.59  & 55.8  & 0.61 \\
     1.4B &  12B  & 1250  & 53.3 & 0.57  & 55.0 & 0.54 \\
\bottomrule
\end{tabular}
\end{sc}
\end{small}
\end{center}
\vskip -0.1in
\end{table}



\appendixsection{Labeled examples}
\label{labeled_examples}

Below, we provide labeled sentences for the 7B/65B LLaMA pairing in the [2, 3) entropy band. Individual tokens are separated by spaces. Tokens in the ``epistemic'' category are colored green. Red tokens are in the second class (``aleatoric-like''). \textit{Note that labels have been class- and token- balanced and that only tokens with small model entropy between 2 and 3 are eligible to be colored here in the first place; green tokens are a very sparse subset of all ``epistemic'' tokens that can be detected by contrasting model entropies.}. Examples are individual sentences containing labels from randomly selected documents in the Wikipedia and Pile (Code) datasets.

Wikipedia:

\begin{itemize}
\item Č T 3 ( Č esk á te lev ize  3 , " T ro j ka ") was the Czech public television channel , operated by Czech Television . Č T 3 broadcast originally in  1 9 9 3 \textcolor{red}{when} it replaced the previous station OK 3 . Un like the other two channels of the Czech Television at the time , Č T 1 and Č T 2 , Č T 3 broadcast its program largely in foreign languages \textcolor{red}{and}  2 4 hours a day .
\item Q ar hd and al - F ard at () is a sub - d istrict \textcolor{green}{located} in \textcolor{green}{Al} Ash ah District , ' Am ran Governor ate , Y emen . Q ar hd and al - F ard at had a population of  2 6 2 3 according to the  2 0 0 4 census .
\item Qu id Pro Qu o is an  1 8 4 4 comedy play by the British writer Catherine G ore , best known for her nov els . It premier ed at the Theatre Royal , Hay market in London on  1 8 June  1 8 4 4 . The original cast included Louis a C ran \textcolor{green}{st} oun N is b ett as Lord Bell am ont , Robert Str ick land as Jer emy Gr ig son , John Buck stone as Captain Si ppet , William Far ren as Sir George M ord ent , Henry How e as R ivers , Julia Ben nett as Lady Mary R ivers , Julia Glo ver as Mrs . Gr ig son , Mrs . Ed win Y arn old as Ellen and Anne Hum by as Br idget Prim .
\item  This ere ct , rh iz om at ous shr ub grows up to tall . Ro ots grow from trailing branches and many short sho ots . The branches are rig id and have a diameter of up to . Second ary \textcolor{red}{branches} develop on the leaf ax ils on the main stem and have a diameter of up to . Bra chy bl asts   ( sh o ots ) grow in the leaf ax ils of the secondary ban ches . These typically grow up to long and secondary bra chy bl asts   are rare . They are white when young .  
 
 \#\# Le aves 
 
 The tri angular leaves grow closely against the branches and are w ool ly \textcolor{red}{on} the upper surface . They are bright green and are slightly in rolled . The leaves growing on the secondary branches are about half the size of those growing on the main st ems .  
\item Arg im und was a Vis ig oth ic us ur per who briefly claimed the king ship in  5 8 9 – 5 9 0 before being put down by the legit imate so ver eign , Re cc ared I .  Following Re cc ared ' s conversion from A rian ism to Catholic ism , a consp i racy , led by Sun na , the A rian bishop of M ér ida , arose to at first place the A rian Seg ga on the throne but failed due to the plot being bet rayed by one of its own named W itter ic . But more u pr is ings followed \& amp ; this one was no different as Arg im und revol ted in  5 8 9 , somewhere in the Kingdom . In response , Re cc ared like with the re bell ion of Seg ga a few years back , sent his General Claud ius to put it down . The Revol t probably only last ed a short few months if not weeks , and Arg im und was captured . Arg im und probably had his hands cut off ( like his prede cess or \textcolor{red}{)} and was ban ished into Ex ile , his further fate afterwards are unknown . 

\end{itemize}

Code:

\begin{itemize}
    \item $\langle$ height $\rangle$ 2 \textcolor{red}{9} 5 $\langle$/ height $\rangle$ 
    
   $\langle$/ rect $\rangle$
   
  $\langle$/ property $\rangle$
  
  $\langle$ widget class =" Q \textcolor{red}{Tab} Widget " name =" tab Widget "$\rangle$
    \item  2 4  3  1 2 \textcolor{red}{1} 7  0  2 0 2  1 6 5  1 6 4 
    
 2 5  3  1 1 3 7  0  2 0 2  1 6 6  1 6 5 
 
 2 6  3  7 9 0  0  2 0 2  1 6 7  1 6 6 
 
 2 7  3  1 1 5 2  0  2 0 2  1 6 8  1 6 7 
 
 2 8  3  1 2 \textcolor{red}{3} 7  0  2 0 2  1 6 9  1 6 8 
 
 2 9  3  2 7 2  0  2 0 2  1 7 0  1 6 9 
 
 3 0  3  6 0 0  0  2 0 2  1 7 1  1 7 0 
 
 3 1  3  1 2 3 0  0  2 0 2  1 7 2  1 7 1 
 
 3 2  3  1 3 2 4  0  2 0 2  1 7 3  1 7 2 
 
 3 3  3  7 3 6  0  2 0 2  1 7 4  1 7 3 
 
 3 4  3  4 6 3  0  2 0 2  1 7 5  1 7 4 
 
 3 5  3  1 \textcolor{red}{2} 3 2  0  2 0 2  1 7 6  1 7 5 
 
 3 6  3  1 4 1 1  0  2 0 2  1 7 7  1 7 6 
 
 3 7  3  2 5 \textcolor{red}{9}  0  2 0 2  1 7 8  1 7 7 
 \item fragment : string 
 
       Fragment  sh ader code 
 
        color : string 
            ' local ', ' shared ' or ' global ' 
        """ 
 
        base \_ d type = [ (' position ', ( np . float 3 2 ,  3 ), ' ! local ', ( 0 ,  0 ,  0 )), 
        
                      (' id ',       ( np . float 3 2 ,  1 ), ' ! local ',  0 ), 
                      
                      (' color ',    ( np . float 3 2 ,  4 ), ' local ', ( 0 ,  0 ,  0 ,  1 )), 
                      
                      (" linewidth ", ( np . float 3 2 ,  1 ), ' global ',  1 ), 
                      
                      (" view port ", ( np . float 3 2 ,  4 ), ' global ', ( 0 ,  0 ,  5 1 2 ,  5 1 2 )) 
                      
                      ] 
 
        dtype = base \_ d type 
        
        if user \_ d type : 
        
            dtype . extend ( user \_ d type )

        if vertex is None : 
        
            vertex = gl sl . get (' collections / raw - path . vert ') 
            
        if transform is None : 
        
            transform = Null Transform () 
            
        self . transform = transform   
        
        if \textcolor{green}{fragment} is None : 
        
            fragment = gl sl . get (' collections / raw - path . f rag ') 
 \item In \textcolor{red}{this} tutorial , we will look at how to connect \textcolor{red}{an} [ ST M 3 2 ][ ST M 3 2 ] board to the K aa platform using [ K \textcolor{green}{aa} Io T Platform Ar duino library ][ K aa Io T Platform Ar duino library ].  You will learn how to create a digital tw in of your device , connect it , send te lem etry and receive commands . 
 \item bool Not ify Wait ers ( const bool f Not ify \textcolor{green}{All} ,
 
                       const Wait Term ination Re ason Term ination Re ason );

\end{itemize}

\appendixsection{Dataset samples}
\label{dataset_samples}

To give a sense of the token distributions and formats of each of our datasets, we provide 5 random samples from each below. Wikipedia and Pile documents are drawn from the respective training sets while examples from the three earmarked Pile subsets are drawn from validation sets. We exclude Pile samples from said subsets. Some longer documents are abridged. Arrows denote linebreaks.

\textsc{Wikipedia} \\
\noindent\rule[1ex]{\textwidth}{1pt}

\begin{lstlisting}
Topsy Chapman (1947 - 2022) was a Jazz and Gospel musician from New Orleans, Louisiana.

## Early life

Chapman was born in Kentwood, Louisiana. She sang and played piano at the age of 3 and was considered a musical prodigy. By the age of 6 or 7, she was earning money performing in churches.

## Career

 Chapman moved to New Orleans when she was 17 where she led her family's Gospel group, The Chapmans. Playwright and director, Vernel Bagneris, recruited the group to appear in his production on 'One Mo' Time'. The show ran at Toulouse Theatre and later premiered Off-Broadway. Chapman performed with her daughters, Yolanda Robinson and Jolynda Phillips, under the group name Topsy Chapman and Solid Harmony. Chapman often performed at New Orleans Jazz Fest and well known in the New Orleans music scene.

## Death

 Chapman died in 2022.
\end{lstlisting}
\noindent\hfil\rule[1ex]{0.8\textwidth}{0.1pt}\hfil
\begin{lstlisting}
The Lyons Crime Family, also known as the Lyons Gang or the Lyons Clan, is a Scottish criminal organisation based in Glasgow. It is one of the most notorious criminal organisations in Scotland, with a long nistory of involvement in organized crime ncluding drug trafficking, extortion, and money laundering. The Lyons Gang is known for its rivalry with another Glasgow-based gang, the Daniel Crime Family (also known as the Daniel Gang or the Daniel Clan), which has resulted in a number of violent incidents over the years.

## Background

The Lyons Crime Family is believed to have been founded in the 1980s by William "Benny" Lyons, who was the leader of the gang at the time until his death in 2006. His brother, Eddie Lyons Sr is also believed to be a prominent member of the organisation. The family's origins can be traced back to the Possilpark area of Glasgow, where the Lyons family had lived for generations. The Lyons are known for their brutal tactics. The exact year when the gang was formed is not clear, but it is believed to have emerged in the early to mid-1980s. At that time, Glasgow was experiencing an increase in drug-related crime and violence, and the Lyons Crime Family saw an opportunity to profit from the illegal drug trade. Over time, the Lyons Crime Family expanded its operations beyond Glasgow and developed links to other criminal organisations, both in Scotland and abroad. Despite the arrests and convictions of some of its members, the gang has remained active and has continued to engage in drug trafficking, extortion, and other illegal activities. Over the years, the Lyons Crime Family has been involved in a number of high-profile crimes, including the murder of Kevin 'Gerbil' Carroll in 2010. The organisation has also been linked to a number of drug trafficking and money laundering operations. Scotland continues to have by far the highest drug death rate recorded by any country in Europe, the Lyons Gang collaborate with the notorious Irish Kinahan Cartel to provide high quality drugs but also to wash the money in the Scottish economy. Despite its criminal activities, the Lyons Crime Family has also been involved in charitable work in the Glasgow area, which has helped to increase its popularity and support among some members of the community. Overall, the Lyons Crime Family is a complex and controversial organisation that has had a significant impact on the criminal landscape in Scotland.
\end{lstlisting}
\noindent\hfil\rule[1ex]{0.8\textwidth}{0.1pt}\hfil
\begin{lstlisting}
Jakez Cornou (1935 - 1 September 2022) was a French historian, ethnologist, and author. He lived in Bigouden and pursued passions in history, ethnography, and heritage of the Bretons. He was a co-founder of the newspapers "" and "Pays de Quimper en Cornouaille". He also directed the publishing house Éditions Sked and was a member of the collective , which sought to promote literature in the "Pays Bigouden".
\end{lstlisting}
\noindent\hfil\rule[1ex]{0.8\textwidth}{0.1pt}\hfil
\begin{lstlisting}
The 2007-08 County Antrim Shield was the 119th edition of the County Antrim Shield, a cup competition in Northern Irish football. Glentoran won the tournament for the 25th time, defeating Crusaders 2-1 in the final.
\end{lstlisting}
\noindent\hfil\rule[1ex]{0.8\textwidth}{0.1pt}\hfil
\begin{lstlisting}
Cyril Aubrey Kemp (12 June 1915 - 25 December 2010) was an Irish tennis player active in the 1930s, 1940s and 1950s. He was also a national representative in the sports of squash and table tennis. The son of an all-round sportsman, Kemp had his best period on the tennis tour in the late 1940s, debuting for the Ireland Davis Cup team in 1946. He was singles runner-up at the Irish championships in 1946 and 1947. His run at the 1947 Irish championships included an upset semi-final win over Tom Brown, who was fresh off making a Wimbledon final. In 1948 he won through to the third round at Wimbledon, before losing to the top seeded Frank Parker.
\end{lstlisting}

\textsc{Pile} \\
\noindent\rule[1ex]{\textwidth}{1pt}
\begin{lstlisting}
I am so sad Charlie one of our beloved cats has cancer out Vet removed a golf ball sized tumor from him today . we are sending it off to be tested and i am praying that she got it all . We will also know what type it is but we fear we will have to make the decision to let him go , he is just 7 years old and i feel it's just too young to have to go :sad: Our other cat George already knows something is wrong because he keeps looking for Charlie. How will i help him cope if we have to put Charlie down ?

chico2

May 18th, 2006, 05:29 PM

I am sorry to hear that,I too had a cat(Peppi)who had a tennis-ball size tumor..I pointed it out to my vet,when the tumor was only quarter-size,she said it was only a fatty deposit.:mad: Peppi was also diabetic..
Weeks later it was huge and Peppi gave up on life and had to be euthanized.
My vet never suggested an operation,so that yours is attempting to remove Charlies could be good news.
George will eventually be ok,but if you loose Charlie:sad:George will miss him very much,another kitten/cat would probably help him.
But lets hope it does not come to that,lets hope Charlie recovers and that it's not as bad as was thought:fingerscr 
\end{lstlisting}
\noindent\hfil\rule[1ex]{0.8\textwidth}{0.1pt}\hfil
\begin{lstlisting}
Listing 136254134. One story floor plan with high vaulted ceilings.

Many upgrades including wood flooring in all living areas, crown molding, and granite kitchen countertops, stainless appliances. much more. Master bedroom has dual sinks in the marble topped vanity and framed mirrors, and walk in closet and access to your private backyard by of sliding glass doors. Close to freeway -> Read More
\end{lstlisting}
\noindent\hfil\rule[1ex]{0.8\textwidth}{0.1pt}\hfil
\begin{lstlisting}
Landscapes of Communism: a History Through Buildings

Owen Hatherley

Allen Lane, 624pp, £25

When I was ten, Mum and Dad took us on our annual family holiday - this time, to Yugoslavia. It was, with hindsight, an especially odd place for them to have taken us. Mum and Dad were certainly not communists but, very much unlike their son, Maggie-loving Tories who'd have voted for a pig wearing a blue rosette. Family holidays had hitherto stuck to old favourites, such as Devon or Somerset. And now here we were, nonchalantly skipping across the Iron Curtain as if popping down to Sainsbury's. Marshal Tito's kind of communism was softer than most, with its open borders and political neutrality. But it was clear from the disappointing lack of tourist tat in the shops that we weren't in Lyme Regis any more, Toto. We were in heaven.
\end{lstlisting}
\noindent\hfil\rule[1ex]{0.8\textwidth}{0.1pt}\hfil
\begin{lstlisting}
Note: Citations are based on reference standards. However, formatting rules can vary widely between applications and fields of interest or study. The specific requirements or preferences of your reviewing publisher, classroom teacher, institution or organization should be applied.

Abstract:

Political theorists consider the challenge of global climate change from a range of perspectives, including conceptual analysis, critical theory, critical legal studies, and neo-Marxism.Read more...
\end{lstlisting}
\noindent\hfil\rule[1ex]{0.8\textwidth}{0.1pt}\hfil
\begin{lstlisting}
Interface

The short video to the left will show you the basic controls and operation of the game.

Please note: the game is in development, so there may be subtle changes between the video and the current state of the game

This will walk you through:

scrolling left and right (you can also scroll up and down if many annotations are made)

making a selection by clicking on the start and ending tokens then clicking annotate

deleting a selection by double clicking

clearly incorrect mentions, reserved mentions

how an agreement between two players is shown

and how to finish a round of the game

Pause and rewind the video if you miss anything.
\end{lstlisting}

\textsc{Pile (Code)} \\
\noindent\rule[1ex]{\textwidth}{1pt}
\begin{lstlisting}
/* iCheck plugin Polaris skin
----------------------------------- */
.icheckbox_polaris,
.iradio_polaris {
    display: block;
    margin: 0;
    padding: 0;
    width: 29px;
    height: 29px;
    background: url(polaris.png) no-repeat;
    border: none;
    cursor: pointer;
}

.icheckbox_polaris {
    background-position: 0 0;
}
    .icheckbox_polaris.hover {
        background-position: -31px 0;
    }
    .icheckbox_polaris.checked {
        background-position: -62px 0;
    }
    .icheckbox_polaris.disabled {
        background-position: -93px 0;
        cursor: default;
    }
    .icheckbox_polaris.checked.disabled {
        background-position: -124px 0;
    }

.iradio_polaris {
    background-position: -155px 0;
}
    .iradio_polaris.hover {
        background-position: -186px 0;
    }
    .iradio_polaris.checked {
        background-position: -217px 0;
    }
    .iradio_polaris.disabled {
        background-position: -248px 0;
        cursor: default;
    }
    .iradio_polaris.checked.disabled {
        background-position: -279px 0;
    }

/* Retina support */
@media only screen and (-webkit-min-device-pixel-ratio: 1.5),
       only screen and (-moz-min-device-pixel-ratio: 1.5),
       only screen and (-o-min-device-pixel-ratio: 3/2),
       only screen and (min-device-pixel-ratio: 1.5) {
    .icheckbox_polaris,
    .iradio_polaris {
        background-image: url(polaris@2x.png);
        -webkit-background-size: 310px 31px;
        background-size: 310px 31px;
    }
}
\end{lstlisting}
\noindent\hfil\rule[1ex]{0.8\textwidth}{0.1pt}\hfil
\begin{lstlisting}
Microsoft Visual Studio Solution File, Format Version 10.00
# Visual Studio 2008
Project("{8BC9CEB8-8B4A-11D0-8D11-00A0C91BC942}") = "Image", "Image\Image_vs90.vcproj", "{DA74060D-73AF-3E8F-A804-FBC960DAC393}"
EndProject
Project("{8BC9CEB8-8B4A-11D0-8D11-00A0C91BC942}") = "Text", "Text\Text_vs90.vcproj", "{0DE18C25-1694-3598-831D-4FA48D113606}"
EndProject
Project("{8BC9CEB8-8B4A-11D0-8D11-00A0C91BC942}") = "Template", "Template\Template_vs90.vcproj", "{27E36FB4-BDAB-3B36-910A-1F1C26853B1E}"
EndProject
\end{lstlisting}
\noindent\hfil\rule[1ex]{0.8\textwidth}{0.1pt}\hfil
\begin{lstlisting}
/* Copyright (C) 2019 Open Information Security Foundation
 *
 * You can copy, redistribute or modify this Program under the terms of
 * the GNU General Public License version 2 as published by the Free
 * Software Foundation.
 *
 * This program is distributed in the hope that it will be useful,
 * but WITHOUT ANY WARRANTY; without even the implied warranty of
 * MERCHANTABILITY or FITNESS FOR A PARTICULAR PURPOSE.  See the
 * GNU General Public License for more details.
 *
 * You should have received a copy of the GNU General Public License
 * version 2 along with this program; if not, write to the Free Software
 * Foundation, Inc., 51 Franklin Street, Fifth Floor, Boston, MA
 * 02110-1301, USA.
 */

/**
 *
 * \author Giuseppe Longo <giuseppe@glongo.it>
 *
 * Implements the sip.stat_msg sticky buffer
 *
 */

#include "suricata-common.h"
#include "threads.h"
#include "debug.h"
#include "decode.h"
#include "detect.h"

#include "detect-parse.h"
#include "detect-engine.h"
#include "detect-engine-mpm.h"
#include "detect-engine-prefilter.h"
#include "detect-content.h"
#include "detect-pcre.h"
#include "detect-urilen.h"

#include "flow.h"
#include "flow-var.h"
#include "flow-util.h"

#include "util-debug.h"
#include "util-unittest.h"
#include "util-unittest-helper.h"
#include "util-spm.h"

#include "app-layer.h"
#include "app-layer-parser.h"

#include "detect-sip-stat-msg.h"
#include "stream-tcp.h"

#include "rust.h"
#include "app-layer-sip.h"

#define KEYWORD_NAME "sip.stat_msg"
#define KEYWORD_DOC  "sip-keywords.html#sip-stat-msg"
#define BUFFER_NAME  "sip.stat_msg"
#define BUFFER_DESC  "sip response status message"
static int g_buffer_id = 0;

static int DetectSipStatMsgSetup(DetectEngineCtx *de_ctx, Signature *s, const char *str)
{
    if (DetectBufferSetActiveList(s, g_buffer_id) < 0)
        return -1;

    if (DetectSignatureSetAppProto(s, ALPROTO_SIP) < 0)
        return -1;

    return 0;
}
\end{lstlisting}
\noindent\hfil\rule[1ex]{0.8\textwidth}{0.1pt}\hfil
\begin{lstlisting}
namespace Alphaleonis.Win32.Vss
{
   /// <summary>The <see cref="VssRestoreType"/> enumeration is used by a requester to indicate the type of restore operation it is about to perform.</summary>
   /// <remarks>
   ///     <para>A requester sets the type of a restore operation using <see cref="IVssBackupComponents.SetRestoreState"/>.</para>
   ///     <!-- <para>A writer can retrieve the type of a restore operation by calling CVssWriter::GetRestoreType.</para> -->
   /// </remarks>
   public enum VssRestoreType
   {
      /// <summary>
      ///       <para>No restore type is defined.</para>
      ///       <para>This indicates an error on the part of the requester.</para>
      /// </summary>
      Undefined = 0,

      /// <summary>The default restore type: A requester restores backed-up data to the original volume from a backup medium.</summary>
      ByCopy = 1,

      /// <summary>
      ///       <para>
      ///               A requester does not copy data from a backup medium, but imports a transportable shadow copy
      ///               and uses this imported volume for operations such as data mining.
      ///       </para>
      ///       <para>
      ///               <b>Windows Server 2003, Standard Edition and Windows Server 2003, Web Edition:</b> This value is not supported. All editions of Windows Server 2003 SP1 support this value.
      ///       </para>
      /// </summary>
      Import = 2,

      /// <summary>A restore type not currently enumerated. This value indicates an application error.</summary>
      Other = 3

   };
}
\end{lstlisting}
\noindent\hfil\rule[1ex]{0.8\textwidth}{0.1pt}\hfil
\begin{lstlisting}
// SPDX-License-Identifier: MIT
// Copyright (c) 2015-2020 Zig Contributors
// This file is part of [zig](https://ziglang.org/), which is MIT licensed.
// The MIT license requires this copyright notice to be included in all copies
// and substantial portions of the software.
// Ported from:
//
// https://github.com/llvm/llvm-project/blob/ 2ffb1b0413efa9a24eb3c49e710e36f92e2cb50b/compiler-rt/lib/builtins/modti3.c

const udivmod = @import("udivmod.zig").udivmod;
const builtin = @import("builtin");
const compiler_rt = @import("../compiler_rt.zig");

pub fn __modti3(a: i128, b: i128) callconv(.C) i128 {
    @setRuntimeSafety(builtin.is_test);

    const s_a = a >> (128 - 1); // s = a < 0 ? -1 : 0
    const s_b = b >> (128 - 1); // s = b < 0 ? -1 : 0

    const an = (a ^ s_a) -% s_a; // negate if s == -1
    const bn = (b ^ s_b) -% s_b; // negate if s == -1

    var r: u128 = undefined;
    _ = udivmod(u128, @bitCast(u128, an), @bitCast(u128, bn), &r);
    return (@bitCast(i128, r) ^ s_a) -% s_a; // negate if s == -1
}
\end{lstlisting}
\textsc{Pile (EuroParl)} \\
\noindent\rule[1ex]{\textwidth}{1pt}
\begin{lstlisting}
Tervetulotoivotukset
Puhemies
(DE) Hyvät parlamentin jäsenet, minulla on suuri ilo toivottaa tervetulleeksi joukko Saksan demokraattisen tasavallan ensimmäisen vapailla vaaleilla valitun parlamentin entisiä jäseniä, jotka istuvat yleisölehterillä.
Euroopan parlamentti pääsi historiankirjoihin päättäessään yhdistää Saksan uudelleen ja hajosi itse pian sen jälkeen. Valtuuskuntaa johtaa Volkskammerin silloinen puheenjohtaja, tohtori Sabine Bergmann-Pohl. Toivotan teidät erittäin lämpimästi tervetulleeksi Euroopan parlamenttiin.
(Suosionosoituksia)
\end{lstlisting}
\noindent\hfil\rule[1ex]{0.8\textwidth}{0.1pt}\hfil
\begin{lstlisting}
4. Informações relativas a medicamentos sujeitos a receita médica (procedimentos comunitários de autorização e de fiscalização de medicamentos) (
Antes da votação da alteração 13:
Christofer Fjellner
Senhor Presidente, tenho uma pequena alteração oral em resultado de um compromisso de última hora entre os grupos políticos relativamente à alteração 13, cujo texto actual, que diz "no prazo de 60 dias a contar da data de recepção da notificação" deve ser alterado para "no prazo de 90 dias a contar da data de recepção da notificação".
Esta alteração foi acordada entre todos os grupos políticos.
\end{lstlisting}
\noindent\hfil\rule[1ex]{0.8\textwidth}{0.1pt}\hfil
\begin{lstlisting}
Übermittlung von Gemeinsamen Standpunkten des Rates: siehe Protokoll
Bogusław Rogalski
(PL) Frau Präsidentin! Entschuldigen Sie bitte, aber ich möchte etwas zu meinem Abstimmungsverhalten bezüglich des Berichts über Mazedonien sagen, wenn das möglich ist.
Die Präsidentin
Herr Rogalski, Sie standen nicht auf der Liste, und ich habe die Stimmerklärungen für abgeschlossen erklärt. Es tut mir leid. Lassen Sie es uns das nächste Mal bitte innerhalb der vorgeschriebenen Frist wissen. Ich weise Sie jedoch darauf hin, dass Sie Ihre Stimmerklärung schriftlich abgeben können.
(Die Sitzung wird um 12.45 Uhr unterbrochen und um 15.00 Uhr wieder aufgenommen.)
\end{lstlisting}
\noindent\hfil\rule[1ex]{0.8\textwidth}{0.1pt}\hfil
\begin{lstlisting}
Velkomst
Formanden
Mine damer og herrer! Jeg byder velkommen til eftermiddagsmødet. Mødet er genoptaget og er det sidste møde før juleferien og nytår. Jeg vil gerne benytte lejligheden til at sige velkommen til direktionen for den tyske kommission for krigsgrave (Volksbund Deutsche Kriegsgräberfürsorge) fra Baden-Württemberg under ledelse af hr. Nothelfer. Med base i Baden-Württemberg står kommissionen for vedligeholdelsen af gravstederne for 90 000 ofre for krig og tyranni i Baden-Württemberg alene. Vi takker Dem for Deres arbejde og byder Dem velkommen til Europa-Parlamentet.
\end{lstlisting}
\noindent\hfil\rule[1ex]{0.8\textwidth}{0.1pt}\hfil
\begin{lstlisting}
4. Särskild rapport från Europeiska ombudsmannen till Europaparlamentet efter förslaget till rekommendation till Europeiska kommissionen avseende klagomål 676/2008/RT (enligt artikel 205.2, 1:a delen, i arbetsordningen) (
\end{lstlisting}

\textsc{Pile (Stack Exchange)} \\
\noindent\rule[1ex]{\textwidth}{1pt}

\begin{lstlisting}
    Q:

window.open location=yes - Firefox 7.01 not showing navigation buttons

I have a window that is created via window.open. When the window is created in Firefox 7.01, I turn on the address bar (location=yes). In Firefox 6.01 - 7.01 the forward and back buttons do not show.  It works perfectly in IE. Can someone tell me/show me how to show navigation buttons using windows.open and location=yes in Firefox?  Thanks!

A:

Try adding toolbar=yes as well.
location=yes is for showing the location bar.
Many browsers will default (or even force) some elements you don't request to show anyway, but you should be explicit about all the elements you wish to show rather than relying on this.
More information is available at MDN.
\end{lstlisting}\
\noindent\hfil\rule[1ex]{0.8\textwidth}{0.1pt}\hfil
\begin{lstlisting}
Q:

Is it ill-advised to make a cover for your thesis in STEM?

I handed in my undergraduate thesis in mathematics and since I had extra time due to COVID-19 lockdown, I had made a cover drawing for it that I included. I've never seen a thesis with a cover in the STEM field before.
Was it a bad decision on my part? Will it look like kowtowing/bootlicking my supervising professor for a better grade? I believe my work was pretty good on its own and I'm afraid I have ruined it with what was intended as a personal touch that I really wanted to add.

A:

It does not matter.
It used to be that theses were printed on paper.  In those days, they had standard covers so they would all look the same when placed on a shelf in the library.
Today nobody cares about a digital cover.
Be prepared to provide a copy without the cover if someone asks for it.

A:

This really comes down to institutional formatting policies. Some may disallow cover art, some may be indifferent to it, and some actually encourage it. For example, my Bachelor's thesis was in a STEM field and included an electron microscopy image on the front cover without any issue. I generally think it's a nice touch, assuming the cover art is relevant to the contents. Further, if that assumption holds it's really difficult to see how it could be interpreted as bootlicking.
I also know of several higher degree theses (Master's, PhD) from the same institution in physics and engineering that had front cover art. And they still print and bind PhD theses (and some Master's theses). They achieve a uniform look on a library shelf since the spine isn't affected. In fact, the institution's website notes that they prefer PhD theses to have similar cover art.
\end{lstlisting}
\noindent\hfil\rule[1ex]{0.8\textwidth}{0.1pt}\hfil
\begin{lstlisting}
Q:

Intelligent spell checking

I'm using NHunspell to check a string for spelling errors like so:
var words = content.Split(' ');
string[] incorrect;
using (var spellChecker = new Hunspell(affixFile, dictionaryFile))
{
    incorrect = words.Where(x => !spellChecker.Spell(x))
        .ToArray();
}

This generally works, but it has some problems. For example, if I'm checking the sentence "This is a (very good) example", it will report "(very" and "good)" as being misspelled. Or if the string contains a time such as "8:30", it will report that as a misspelled word. It also has problems with commas, etc.
Microsoft Word is smart enough to recognize a time, fraction, or comma-delimited list of words. It knows when not to use an English dictionary, and it knows when to ignore symbols. How can I get a similar, more intelligent spell check in my software? Are there any libraries that provide a little more intelligence?
EDIT:
I don't want to force users to have Microsoft Word installed on their machine, so using COM interop is not an option.

A:

If your spell checker is really that stupid, you should pre-tokenize its input to get the words out and feed those one at a time (or as a string joined with spaces). I'm not familiar with C#/.NET, but in Python, you'd use a simple RE like \w+ for that:
>>> s = "This is a (very good) example"
>>> re.findall(r"\w+", s)
['This', 'is', 'a', 'very', 'good', 'example']

and I bet .NET has something very similar. In fact, according to the .NET docs, \w is supported, so you just have to find out how re.findall is called there.
\end{lstlisting}
\noindent\hfil\rule[1ex]{0.8\textwidth}{0.1pt}\hfil
\begin{lstlisting}
Q:

Create list of variable type

I am trying to create a list of a certain type.
I want to use the List notation but all I know is a "System.Type"
The type a have is variable. How can I create a list of a variable type?
I want something similar to this code.
public IList createListOfMyType(Type myType)
{
     return new List<myType>();
}

A:

Something like this should work.
public IList createList(Type myType)
{
    Type genericListType = typeof(List<>).MakeGenericType(myType);
    return (IList)Activator.CreateInstance(genericListType);
}

A:

You could use Reflections, here is a sample:
    Type mytype = typeof (int);

    Type listGenericType = typeof (List<>);

    Type list = listGenericType.MakeGenericType(mytype);

    ConstructorInfo ci = list.GetConstructor(new Type[] {});

    List<int> listInt = (List<int>)ci.Invoke(new object[] {});
\end{lstlisting}
\noindent\hfil\rule[1ex]{0.8\textwidth}{0.1pt}\hfil
\begin{lstlisting}
Q:

Json data containing NaN results in error during ajax request

I an endpoint exposed via web api which spews out JSON like this:
"[{"SomeId":1,"SomeName":"Some name 1","Parameter1":1.13,"Parameter2":3.0 ...

to jquery ajax get requests. Everything works fine until the JSON contains NaN (not a number):
"[{"SomeId":1,"SomeName":"Some name 1","Parameter1":1.13,"Parameter2":NaN ...

If this type of data is returned the error callback is hit.
Is this a known issue? Is jquery or web api to blame? How can I mitigate this? Thanks.

A:

This happens because the token NaN is invalid in JSON. JSON is a subset of JavaScript's object initializer syntax, which doesn't include several aspects of it, including NaN (and Infinity, undefined, ...).
The web API providing that output is producing invalid JSON, so any conforming JSON parser will complain about it. You'll need to fix (or ask the providers of it to fix) the API.
\end{lstlisting}

\end{document}